\documentclass[sigconf]{acmart}
\makeatletter
\def\@ACM@checkaffil{
    \if@ACM@instpresent\else
    \ClassWarningNoLine{\@classname}{No institution present for an affiliation}%
    \fi
    \if@ACM@citypresent\else
    \ClassWarningNoLine{\@classname}{No city present for an affiliation}%
    \fi
    \if@ACM@countrypresent\else
        \ClassWarningNoLine{\@classname}{No country present for an affiliation}%
    \fi
}
\makeatother
\copyrightyear{2023}
\acmYear{2023}
\setcopyright{acmlicensed}\acmConference[MM '23]{Proceedings of the 31st ACM International Conference on Multimedia}{October 29-November 3, 2023}{Ottawa, ON, Canada}
\acmBooktitle{Proceedings of the 31st ACM International Conference on Multimedia (MM '23), October 29-November 3, 2023, Ottawa, ON, Canada}
\acmPrice{15.00}
\acmDOI{10.1145/3581783.3611971}
\acmISBN{979-8-4007-0108-5/23/10}
\usepackage{times}
\usepackage{epsfig}
\usepackage{amsthm}
\usepackage{subfigure}
\usepackage{fancybox}
\usepackage{tikz}
\usepackage{environ}
\usepackage{algorithm}
\usepackage{algorithmic}

\usepackage{cases}
\usepackage{tgpagella}
\usepackage{framed,multirow}
\usepackage{latexsym}
\usepackage{bm}
\usepackage{textcomp}
\usepackage{hyperref}
\usepackage{soul}
\usepackage{url}
\usepackage[utf8]{inputenc}
\usepackage{graphicx}
\usepackage{amsmath}
\usepackage{booktabs}
\usepackage{helvet}
\usepackage{courier}
\usepackage{amsfonts}
\usepackage{verbatim}

\usepackage{amssymb}
\usepackage{subfigure}
\usepackage{setspace}
\usepackage{epstopdf}
\usepackage{color}
\usepackage{leftidx}
\usepackage{multicol}
\usepackage{lineno}
\usepackage{threeparttable}
\usepackage{multirow}
\usepackage{rotating}
\usepackage{makecell}
\usepackage{cases}
\usepackage{setspace}
\usepackage{ulem}
\newcommand{\eat}[1]{}
\newcommand{\ie}{{\it{i.e.,~}}}
\newcommand{\eg}{{\it{e.g.,~}}}

\newcommand{\etal}{{\it{ et al.~}}}

\usepackage{todonotes}

\definecolor{p2}{RGB}{121,64,255}

\newcommand{\ours}{{CoCoMG}}

\usepackage{fancyhdr}
\AtBeginDocument{%
  \providecommand\BibTeX{{%
    \normalfont B\kern-0.5em{\scshape i\kern-0.25em b}\kern-0.8em\TeX}}}

\begin{document}

\title{Unsupervised Multiplex Graph Learning\\ with Complementary and Consistent Information}

\author{Liang Peng}
\affiliation{%
 \institution{School of Computer Science and Engineering, University of Electronic Science and Technology of China, Chengdu, China}
 }

 \author{Xin Wang}
\affiliation{%
 \institution{School of Computer Science and Engineering, University of Electronic Science and Technology of China, Chengdu, China}
 }

 \author{Xiaofeng Zhu}
 \authornote{Corresponding author (seanzhuxf@gmail.com). \\ 
This work was supported in part by 
National Key Research and Development Program of China under Grant 2022YFA1004100,  
the Medico-Engineering Cooperation Funds from University of Electronic Science and Technology of China under Grant  ZYGX2022YGRH009 and Grant ZYGX2022YGRH014, 
and the Guangxi ``Bagui'' Teams for Innovation and Research, China.}
\affiliation{
\institution{School of Computer Science and Engineering, University of Electronic Science and Technology of China, Chengdu, China}~\\
Shenzhen Institute for Advanced Study, University of Electronic Science and Technology of China, Shenzhen, China
 }

\def\authors{Liang Peng, Xin Wang, Xiaofeng Zhu}

\renewcommand{\shortauthors}{Liang Peng et al.}

\begin{abstract}
Unsupervised multiplex graph learning (UMGL) has been shown to achieve significant effectiveness for different downstream tasks by exploring both complementary information and consistent information among multiple graphs. However, previous methods usually overlook the issues in practical applications, \ie the out-of-sample issue and the noise issue.
To address the above issues, in this paper, we propose an effective and efficient UMGL method to explore both  complementary and consistent information. To do this, our method employs multiple MLP encoders rather than graph convolutional network (GCN) to conduct representation learning with two constraints, \ie preserving the local graph structure among nodes to handle the out-of-sample issue, and maximizing the correlation of multiple node representations to handle the noise issue. 
Comprehensive experiments demonstrate that our proposed method achieves superior effectiveness and efficiency over the comparison methods and effectively tackles those two issues.
Code is available at https://github.com/LarryUESTC/CoCoMG.
\end{abstract}

\maketitle
\section{Introduction} \label{sec_intro}
Multiplex graph is a type of data that consists of multiple graphs, where each graph represents a different type of relationship among nodes \cite{battiston2017new, gan2022multi, liu2021graph}. This kind of data is prevalent in various practical applications, such as social networks and biological networks \cite{zhang2019multiplex, ZhuMa2022, yuan2021adaptive, li2021multiplex}. Recently, there has been  growing interest in unsupervised multiplex graph learning (UMGL) due to its ability to extract valuable information from multiple relationships among nodes without relying on label information \cite{DMGIParkK0Y20, ChuFYZHB19,zhang2022unsupervised, WANG2022}.

The success of existing UMGL methods lies in that they are able to explore both  \textit{complementary} information and  \textit{consistent} information among multiple graphs, so that UMGL  can provide more comprehensive information than graph learning on the single graph \cite{ChuFYZHB19, zhou2020multiple,xie2023Self}. The complementary information refers to different types of relationships that can supply each other and provide a more comprehensive understanding of the hidden patterns or structures in the data \cite{cen2019representation}. For example, friendships and communication relationships provide complementary information to identify individuals belonging to the same social circle \cite{wang2022rethinking}. The consistent information means consistent relationships among nodes or structures across multiple graphs, which can help the model recognize the same category effectively. For example, the consistency of gene expression and protein-protein interactions is helpful to identify categories in biological networks \cite{zhu2022structure,li2021multiplex}. 
Obviously, either complementary information or consistent information often plays an important role in UMGL. 

Various methods have been proposed in the literature \cite{DMGIParkK0Y20, WangLHS21,jing2021hdmi,zhu2022structure} to explore and leverage either complementary information or consistent information. For instance,  Jing \etal apply graph convolutional networks (GCN) \cite{kipf2016variational} to first obtain representations from each graph and then to fuse these representations by an attention mechanism \cite{NIPS2017_3f5ee243}, aiming at  exploring complementary information among multiple graphs \cite{jing2021hdmi}. However, its effectiveness is limited by the noisy information as it ignores the intrinsic correlation (\ie consistent information) across different graphs. To address this limitation, recent researchers \cite{DMGIParkK0Y20, WangLHS21, zhu2022structure, jing2022x} have explored contrastive learning techniques \cite{oord2018representation} to simultaneously explore complementary and consistent information for UMGL. For example, to explore the consistent information, Park \etal conduct contrastive learning between node representations and graph representations \cite{DMGIParkK0Y20, jing2021hdmi} and Zhu \etal conduct contrastive learning among node representations from different graphs \cite{zhu2022structure}. Thus, it is crucial  to learn both complementary information and consistent information for conducting UMGL.

\begin{figure*}[!t]
\begin{center}
\vspace{-3mm}
\includegraphics[width=1.0\textwidth,trim =300 450 260 200,clip]{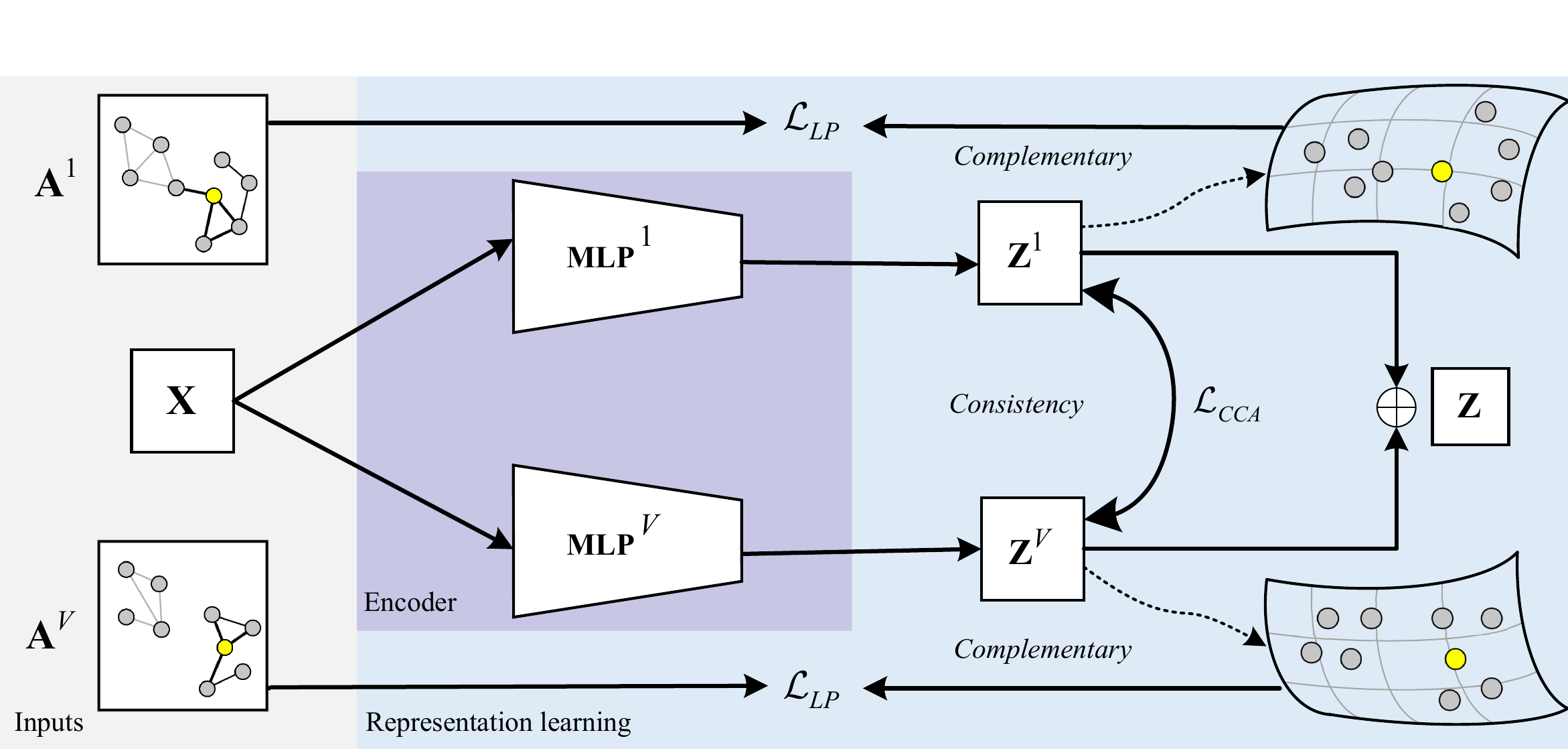}
\caption{The flowchart of the proposed \ours{}.
Given the original node feature matrix $\mathbf{X}$, \ours{} employs MLP encoders with two constraints ($\mathcal{L}_{LP}$ and $\mathcal{L}_{CCA}$) to generate representation $\mathbf{Z}^{v}$ ($v = 1, ..., V$).
Specifically, the local preserve loss $\mathcal{L}_{LP}$ is designed to extract the complementary information of every graph and deal with the  out-of-sample issue. The canonical correlation analysis (CCA) loss $\mathcal{L}_{CCA}$ is designed to extract the consistent information among all graphs and deal with the noise issue.
Finally, the final representation $\mathbf{Z}$ is obtained for downstream tasks by  averagely fusing all representations.
}
\label{fig_flowchart}
\end{center}
\end{figure*}


Despite the important advances, previous UMGL methods for simultaneously considering complementary and consistent information usually overlook some issues, such as the  out-of-sample issue \cite{ma2018depthlgp} and the noise issue \cite{xu2013survey}, resulting in a gap between the existing methods and their practical applications. On the one hand,  previous UMGL methods \cite{DMGIParkK0Y20, WangLHS21,jing2021hdmi,zhu2022structure,han2023CoLM2S} can not directly infer representations for unseen nodes because they do not directly generate a prediction model to predict unseen nodes. 
Specifically, the key issue for either the prediction model generation or the UMGL is the message-passing mechanism, which needs to aggregate information from nodes. In real applications, aggregating information  may be efficient for the single graph, but it is usually inefficient for multiple graphs \cite{velivckovic2022message}. 
On the other hand, previous methods cannot effectively and efficiently deal with noise for  UMGL. For example, previous UMGL methods \cite{zhu2021graph,DMGIParkK0Y20, WangLHS21, zhu2022structure,ZHANG2023Self} investigate contrastive learning techniques to explore consistent information,  but noise (\eg incorrect connections) may be aggregated into the process of representation learning, resulting in ineffective UMGL. Additionally, many previous methods require constructing and embedding positive and negative pairs for every graph, which is computationally inefficient. Overall, it is challenging to effectively and efficiently explore both complementary and consistent information for UMGL.

To overcome the above issues, in this paper, we explore complementary and consistent information in a unified framework for UMGL by proposing an effective and efficient UMGL method \ours{}, \ie \textit{learning \underline{Co}mplementary and \underline{Co}nsistent information for \underline{M}ultiplex \underline{G}raph in practical scenarios}, as shown in Figure \ref{fig_flowchart}. Firstly, we explore the complementary information for UMGL by applying the Multi-Layer Perceptron (MLP) encoders and designing a local preserve objective function to achieve effectiveness and efficiency, as well as to tackle the out-of-sample issue. 
Specifically, the representations generated by the MLP encoders are with low computational cost (\ie efficiency), and the representations are expressive enough to characterize the graph structure (\ie effectiveness). Moreover, the MLP encoders implicitly learn the mappings from the features to the graph structure, enabling to directly predict the embedding of unseen nodes only with their features.
Secondly, we explore the consistent information by conducting an objective function that maximizes the correlation of multiple representations of nodes (\ie effectiveness) without constructing negative pairs (\ie efficiency). Meanwhile, the consistent information extraction can also tackle the noise issue as it can balance the local preserve objective function. 
As a result, \ours{} can effectively and efficiently explore both complementary and consistent information to deal with the out-of-sample issue and the noise issue. 

Compared to previous UMGL methods, the main contributions of our method can be summarized as follows:
\begin{itemize}
    \item The proposed method can tackle the out-of-sample issue and the noise issue for UMGL. This could reduce the gap between the UMGL methods and their practical applications.
    \item The proposed method achieves effectiveness and efficiency for exploring both complementary and consistent information for UMGL.
    Comprehensive experiments verified them by comparing \ours{} with comparison methods on real datasets.
\end{itemize}



\section{Methodology} \label{sec_method}
\textbf{Notations.} 
Let $\left \{\mathbf{A}^{v} \in \mathbb{R} ^{N \times N} \right \}^{V}_{v = 1} $ be a set of graphs, where $N$ is the number of node samples and $V$ is the number of graphs. 
$\mathbf{X} \in \mathbb{R} ^{N \times D}$ denotes original node features, where $D$ is the dimension of node features. 
The goal of the UMGL methods is to learn low-dimensional representations $\mathbf{Z}\in \mathbb{R} ^{N \times d}$ for various downstream tasks, where $d$ is the dimension of the learned representations, and $d \ll D$.

\subsection{Motivation} \label{sec_motivation}
Although previous UMGL methods have demonstrated the importance of extracting complementary and consistent information, a few of them have considered the issues in practical scenarios, such as the  out-of-sample issue and the noise issue. 

First, the existing UMGL methods \cite{DMGIParkK0Y20, jing2021hdmi, peng2022Reverse, pan2021multi} do not generate a specific representation model to directly infer representations for unseen nodes, which is known as \textit{the out-of-sample issue}. 
Specifically, previous UMGL methods heavily relied on the message-passing mechanism encoder (\eg GCN) to obtain node representations, which use multiple graphs to aggregate hidden representations from neighbors in each layer. However, such aggregation from seen nodes to unseen nodes requires reconstructing multiple graphs, which is time-consuming and complex in practical applications. Although some methods \cite{hamilton2017inductive, velickovic2018graph, ma2018depthlgp} have touched this issue by avoiding reconstructing the graph within the single graph, they still rely on neighbor information aggregation for inferring the representations of unseen nodes, which leads to inefficiency in UMGL. 

Second, in practical applications, the graph structure usually contains noisy edges, which can be even more severe in UMGL with multiple graphs. This results in \textit{the noise issue}. Although previous methods \cite{DMGIParkK0Y20, jing2021hdmi, WangLHS21} have considered using contrastive learning to explore consistent information to resist noisy information, their effectiveness  might significantly degenerate for dealing with noisy graphs (verified in Section \ref{sec_exp_robust}).  The main reason is possible that  contrastive learning involves aggregating representations of positive pairs and separating representations of negative pairs. This process necessitates designing and constructing suitable pairs of positive and negative pairs to achieve effectiveness. However, whether it is searching for suitable positive and negative pairs or representing them, both can be impacted  by noisy information, resulting in ineffectiveness in practical applications. Moreover, obtaining the representations of positive and negative pairs can also be inefficient in multiple graph learning.

To address the aforementioned  issues, we propose an efficient and effective UMGL method (\ie \ours{}) with the framework presented in Figure \ref{fig_flowchart}. \ours{} extracts the complementary information for every graph (\ie Section \ref{sec_method_comple}) and the consistent information among all graphs (\ie Section \ref{sec_method_consis}). 

\subsection{Complementary information extraction} \label{sec_method_comple}
The above analysis implies that previous methods predominantly rely on the message-passing mechanism to explore complementary information, without adequately considering their applicability to UMGL. To address this limitation, one of  the straightforward methods is to use the MLP encoder instead of the message-passing mechanism. However, the MLP encoder is hard to capture local structures and relationships among nodes, easily leading to inferior performance.

In this paper, we introduce an efficient and effective method to  explore the complementary information for every graph based on the following three conditions: (i) \textit{Deep.} Deep learning models aim at stacking multiple layers to gradually extract complex and high-level features \cite{lecun2015deep}; (ii) \textit{High-order proximity.} Multi-hop relationships among nodes have been demonstrated to  provide  comprehensive relationships among nodes \cite{zhu2018high, jing2021hdmi}; (iii) \textit{Smooth.} Either traditional graph learning methods (\eg shallow learning) or deep graph learning methods (\eg GCN) have been demonstrated to preserve the local structure among nodes for achieving effectiveness \cite{dong2016learning, kipf2017semisupervised}.

\textit{Deep.}  Given  $\mathbf{X}$, we first employ the MLP to extract the representation for nodes in every graph as follows:
\begin{equation}
\mathbf{Z}^{v} = \mathcal{F}_{\Theta^{v}}^{v}(\mathbf{X}),
\label{eq_Z}
\end{equation}
where $\mathbf{Z}^{v}$ is the representations matrix of nodes in the $v$-th graph obtained by the MLP encoder $\mathcal{F}_{\Theta^{v}}^{v}$ with the parameters $\Theta^{v}$. The parameters of the MLP encoders are not shared across different graphs  as every MLP encoder  has unique sets of parameters that are optimized for the specific graph.

\textit{High-order proximity.} We define an adjacency matrix to capture the relationship between nodes and their multi-hop neighbors (\eg two-hop and three-hop neighbors),  denoted by $\mathbf{W}^{v}= \mathbf{A}^{v}(\mathbf{A}^{v})^{T}$ with two-hop neighbors, which results in a more informative representation of nodes. In contrast, the original adjacency matrix  $\mathbf{A}^{v}$ represents the relationship (\eg similarity) between nodes and their one-hop neighbors, and thus  is usually sparse and often lacks information. 

\textit{Smooth.} We encourage neighbor nodes close as much as possible in the embedding space, and \ours{} designs the following objective function to preserve the local structure:
\begin{equation}
\mathcal{L}_{LP} = -\sum_{v=1}^{V}\sum_{i=1}^{N}log\left(\mathbf{w}_{ij}^{v}\overline{\mathcal{S}}(\mathbf{z}_{i}^{v},\mathbf{z}_{j}^{v}) \right),
\label{eq_loss_LP}
\end{equation}
where $\mathbf{w}^{v}_{ij} \in \mathbf{W}^{v}$ is the strength or similarity between the $i$-th node and its multi-hop neighbor, $\overline{\mathcal{S}}(\mathbf{z}_{i}^{v},\mathbf{z}_{j}^{v})$ is the normalized value of the similarity score $\mathcal{S}(\mathbf{z}_{i}^{v},\mathbf{z}_{j}^{v}):= exp({cos(\mathbf{z}_{i}^{v},\mathbf{z}_{j}^{v})})$, and  $exp()$ is involved to scale the value of the cosine similarity. 

\textit{Tackling the out-of-sample issue.} 
To infer the representation for unseen nodes, previous UMGL methods need to reconstruct multiple graphs \cite{jing2021hdmi, WangLHS21, wang2022collaborative, lee2022relational} or retrain the whole model \cite{DMGIParkK0Y20, pan2021multi}, which is time-consuming and complex in practical applications. To address this issue, the proposed method only uses the original features of the unseen nodes (\ie $\textbf{X}_{un}$) to infer their representations, which is efficient and simple. Specifically, after the whole network (\ie $\{\mathcal{F}_{\Theta^{v}}^{v}\}_{v=1}^{V}$) is trained, we directly apply each MLP encoder to obtain the representations for unseen nodes as follows:
\begin{equation}
\mathbf{Z}^{v}_{un} = \mathcal{F}_{\Theta^{v}}^{v}(\mathbf{X}_{un}),
\label{eq_Z_out}
\end{equation}
where the representations will be fused by the average operation to obtain the final representations for unseen nodes. In this way, we only use the original feature information in the inference process, and our method achieves efficient and simple inference for unseen nodes without  constructing new matrices or retraining. Despite its efficiency, our method is also able to obtain effective representations for unseen nodes because it can implicitly learn the mapping from features to structures by Eq. (\ref{eq_loss_LP}), which can preserve the local structure among the training nodes and unseen nodes.

The advantages of our proposed complementary information extraction are listed as follows. First, we employ the MLP as encoders instead of  GCN, which makes representation learning more efficient. Second, the representations characterize the graph structure by satisfying the aforementioned three conditions, making them effective. Third, the MLP encoders implicitly learn the mappings from the features to graph structure, enabling them to solve the out-of-sample issue. Consequently, \ours{} is effective and efficient in exploring complementary information while also addressing the out-of-sample issue. 

\subsection{Consistent information extraction} \label{sec_method_consis}
Although node representations on multiple graphs provide complementary information, they can also contain noisy information that decreases the model's robustness. To reduce the impact of noise, recent studies apply contrastive learning to extract the consistent information across multiple graphs, but they easily result in inefficiency and ineffectiveness. 

In this paper, we propose to effectively and efficiently exploit the consistent information by directly maximizing the correlation among the representations of each graph based on the following conditions: (i) \textit{Correlation.} Enforcing consistent node representations across different graphs, as canonical correlation analysis (\ie CCA) does \cite{hardoon2004canonical}; (ii) \textit{Decorrelation.} Enforcing the disagreement of node representation in each graph to maintain discriminative characteristics \cite{jing2021understanding}. 

To do this, given the representation matrices $\mathbf{Z}^{v}$ ($v = 1, ..., V$), we maximizes the correlation coefficient $\rho$ between two representation matrices $\mathbf{Z}^{a}$ and $\mathbf{Z}^{b}$, \ie
\begin{equation}
\rho\left(\mathbf{Z}^{a}, \mathbf{Z}^{b}\right)=\frac{ (\mathbf{Z}^{a})^{\top}\mathbf{Z}^{b}}{\sqrt{\left((\mathbf{Z}^{a})^{\top}\mathbf{Z}^{a}\right)\left((\mathbf{Z}^{b})^{\top}\mathbf{Z}^{b}\right)}}.
\label{eq_opt_CCA}
\end{equation}

After replacing the variance item  $(\mathbf{Z}^{a})^{\top}\mathbf{Z}^{a}$ with the constrain $\Sigma_{\mathbf{Z}^{a},\mathbf{Z}^{a}} = \Sigma_{\mathbf{Z}^{b},\mathbf{Z}^{b}}= \mathbf{I}$, we  define the CCA regularization loss for UMGL as:
\begin{equation}
\mathcal{L}_{\text{CCA}}=- \sum_{a=1}^{V}\sum_{b>a}^{V}\mathbf{Z}^{a}(\mathbf{Z}^{b})^{\top}+\gamma \sum_{v=1}^{V}\left\|\left(\mathbf{Z}^{v}\right)^{\top} \mathbf{Z}^{v}-\mathbf{I}\right\|_{F}^{2},
\label{eq_loss_LCCA}
\end{equation}
where $\gamma$ is the trade-off coefficient. In Eq. (\ref{eq_loss_LCCA}), the first term encourages the \textit{correlation} among node representations across different graphs. The second term maintains the \textit{decorrelation} among node representations in each graph, which ensures that individual dimensions of the learned representation are uncorrelated to avoid trivial solutions.

\textit{Tackling the noise issue.} One of the challenges for UMGL is to deal with noisy structures in multiplex graph, where the qualities of different graphs captured by various sensors cannot be guaranteed, \eg the graph structures from some sensors may be unreliable and bring noisy information. Previous UMGL methods \cite{DMGIParkK0Y20, jing2021hdmi, WangLHS21, pan2021multi, lee2022relational} first use GCN on each graph to obtain multiple representations and then apply contrastive learning to maintain the consistency among multiple representations for noise resistance, which is both ineffective and inefficient. Differently, the proposed method first obtains multiple representations by the MLP encoders, which means the noisy information will not be directly introduced into multiple representations. Note that, previous UMGL methods need to aggregate representation from neighborhoods, which will directly incorporate  the noisy information into multiple representations.  We then apply the CCA loss Eq. (\ref{eq_loss_LCCA}) to directly maintain the consistency among multiple representations without negative pairs, which can resist the noisy influence from noisy graphs by balancing the local preserve loss in Eq. (\ref{eq_loss_LP}). Specifically,  if the view-specific representation preserves the local structure of the noisy graph, \ie minimizing Eq. (\ref{eq_loss_LP}),  the obtained representations are inconsistent across multiple graphs. As a result, the learnable model will be heavily penalized by Eq. (\ref{eq_loss_LCCA}).

The advantages of our proposed consistent information extraction are listed as follows. First,  the proposed method can effectively deal with the noise issue, as the noisy information is not directly incorporated into the representations. Moreover, the noisy information in Eq. (\ref{eq_loss_LP}) can be corrected and reduced by Eq. (\ref{eq_loss_LCCA}). Second, the proposed method is more efficient than previous methods, as it does not need to construct positive and negative pairs (verified in Section \ref{sec_exp_robust}). 

\subsection{Loss function}
Integrating  Eq. (\ref{eq_loss_LP}) with  Eq. (\ref{eq_loss_LCCA}), the final objective function of \ours{} is:
\begin{equation}
\mathcal{L}=  \mathcal{L}_{\text{LP}} + \beta \mathcal{L}_{\text{CCA}},
\label{eq_loss}
\end{equation}
where $\beta$ achieves the trade-off between $\mathcal{L}_{\text{LP}}$ and $\mathcal{L}_{\text{CCA}}$. In the optimization of \ours{}, $\mathcal{L}_{\text{LP}}$ explores the complementary information  for every graph, and $\mathcal{L}_{\text{CCA}}$ explores the consistent information across multiple graphs. Subsequently,  we fuse the representation for multiple graphs by the average operation  to obtain the final representation (\ie $\mathbf{Z} = \frac{1}{V} \sum_{v=1}^{V} \mathbf{Z}^{v}$), which is further used for downstream tasks.

\begin{table*}[ht]
\centering
\caption{Classification performance (\ie Macro-F1 and Micro-F1) of all methods on all multiplex graph datasets.}
\begin{tabular}{lcccccccc}
\toprule
\multirow{2}{*}{\textbf{Method}}&\multicolumn{2}{c}{\textbf{ACM}}& \multicolumn{2}{c}{\textbf{IMDB}}& \multicolumn{2}{c}{\textbf{DBLP}}& \multicolumn{2}{c}{\textbf{Freebase}} \\
\cmidrule(r){2-3} \cmidrule(r){4-5} \cmidrule(r){6-7} \cmidrule(r){8-9}
&Macro-F1&Micro-F1&Macro-F1&Micro-F1&Macro-F1&Micro-F1 &Macro-F1&Micro-F1\\
\midrule
Deep Walk  &73.9    $\pm$  0.2&74.8   $\pm$ 0.2 &42.5   $\pm$ 0.2&43.3   $\pm$ 0.4&88.1   $\pm$ 0.2&89.5   $\pm$ 0.3&49.3   $\pm$ 0.3&52.1   $\pm$ 0.2 \\
GCN  &86.9   $\pm$ 0.3&87.0   $\pm$ 0.2&45.7   $\pm$ 0.4&49.8   $\pm$ 0.2&90.2   $\pm$ 0.2&90.9   $\pm$ 0.2&50.5   $\pm$ 0.2&53.3   $\pm$ 0.2 \\
GAT  &85.0   $\pm$ 0.2&84.9   $\pm$ 0.1&49.4   $\pm$ 0.2&53.6   $\pm$ 0.4&91.0   $\pm$ 0.4&92.1   $\pm$ 0.2&55.1   $\pm$ 0.3&59.7   $\pm$ 0.4\\
DGI   &88.1   $\pm$ 0.5&88.1   $\pm$ 0.4&45.1   $\pm$ 0.2&46.7   $\pm$ 0.2&90.3   $\pm$ 0.1&91.1   $\pm$ 0.4&54.9   $\pm$ 0.1&58.2   $\pm$ 0.2 \\
\midrule
MNE&79.2   $\pm$ 0.4&79.7   $\pm$ 0.2&44.7   $\pm$ 0.5&45.6   $\pm$ 0.3&89.3   $\pm$ 0.2&90.6   $\pm$ 0.4&52.1   $\pm$ 0.3&54.3   $\pm$ 0.2 \\
%
DMGI &89.8   $\pm$ 0.2&89.8   $\pm$ 0.2&52.2   $\pm$ 0.2&53.7   $\pm$ 0.3&\textbf{92.1   $\pm$ 0.2}&\textbf{92.9   $\pm$ 0.3}&54.9   $\pm$ 0.1&57.6   $\pm$ 0.2 \\
DMGIattn &88.7   $\pm$ 0.2&88.7   $\pm$ 0.2&52.6   $\pm$ 0.2&53.6   $\pm$ 0.4&90.9   $\pm$ 0.2&91.8   $\pm$ 0.3&55.8   $\pm$ 0.4&58.3   $\pm$ 0.1 \\
HDMI &90.1   $\pm$ 0.3&90.1   $\pm$ 0.1&55.6   $\pm$ 0.3&57.3   $\pm$ 0.3&91.3   $\pm$ 0.2&92.2  $\pm$ 0.5&56.1   $\pm$ 0.2&59.2   $\pm$ 0.2\\
HeCo &88.2   $\pm$ 0.2&88.3   $\pm$ 0.3&50.8   $\pm$ 0.3&51.7   $\pm$ 0.3&91.0   $\pm$ 0.3&91.6  $\pm$ 0.3&59.2   $\pm$ 0.3&61.7   $\pm$ 0.2\\
MCGC &90.2   $\pm$ 0.8&90.0   $\pm$ 0.1&56.3   $\pm$ 0.5&57.5   $\pm$ 0.6&91.9   $\pm$ 0.3&92.1   $\pm$ 0.4&56.6   $\pm$ 0.1&59.4   $\pm$ 0.1 \\
CKD &90.4   $\pm$ 0.3&90.5   $\pm$ 0.2&54.8   $\pm$ 0.2&57.7   $\pm$ 0.3&92.0   $\pm$ 0.2&92.3   $\pm$ 0.5&60.4   $\pm$ 0.4&62.9   $\pm$ 0.4 \\
RGRL&90.3   $\pm$ 0.4&90.2   $\pm$ 0.4&    52.1 $\pm$ 0.2  & 55.5 $\pm$ 0.5  &   91.7 $\pm$ 0.2 &  92.0  $\pm$ 0.3 &   59.4 $\pm$ 0.4 &  62.1 $\pm$ 0.3  \\
\textbf{CoCoMG} &\textbf{92.8   $\pm$ 0.1}&\textbf{92.7   $\pm$ 0.1 }&\textbf{58.3   $\pm$ 0.4} &\textbf{59.6   $\pm$ 0.4  }&\textbf{92.1  $\pm$ 0.1}&\textbf{92.9   $\pm$ 0.1}&\textbf{60.8   $\pm$ 0.1}&\textbf{63.8   $\pm$ 0.9}\\

\bottomrule
\end{tabular}
\label{tabnode}
\end{table*}

\begin{table*}[ht]
\centering
\caption{Clustering performance (\ie Accuracy and NMI) of unsupervised methods on all multiplex graph datasets.}
\begin{tabular}{lccccccccccccc}
\toprule
\multirow{2}{*}{\textbf{Method}}&\multicolumn{2}{c}{\textbf{ACM}}& \multicolumn{2}{c}{\textbf{IMDB}}& \multicolumn{2}{c}{\textbf{DBLP}}& \multicolumn{2}{c}{\textbf{Freebase}} \\
\cmidrule(r){2-3} \cmidrule(r){4-5} \cmidrule(r){6-7} \cmidrule(r){8-9}
&Accuracy&NMI&Accuracy&NMI&Accuracy&NMI&Accuracy&NMI\\
\midrule

Deep Walk  & 64.5 $\pm$ 0.7 & 41.6 $\pm$ 0.5 & 42.1 $\pm$ 0.4 & 1.5 $\pm$ 0.1 & 89.5 $\pm$ 0.4 & 69.0 $\pm$ 0.2 & 44.5 $\pm$ 0.6 & 12.8 $\pm$ 0.4 \\
DGI   & 81.1 $\pm$ 0.6 & 64.0 $\pm$ 0.4 & 48.9 $\pm$ 0.2 & 8.3 $\pm$ 0.3 & 85.4 $\pm$ 0.3 & 65.6 $\pm$ 0.4 & 52.9 $\pm$ 0.2 & 17.8 $\pm$ 0.2 \\
\midrule

MNE & 69.1 $\pm$ 0.2 & 54.5 $\pm$ 0.3 & 46.5 $\pm$ 0.3 & 4.6 $\pm$ 0.2 & 86.3 $\pm$ 0.3 & 68.4 $\pm$ 0.2 & 45.1 $\pm$ 0.5 & 13.3 $\pm$ 0.7 \\
DMGI & 88.4 $\pm$ 0.3 & 68.7 $\pm$ 0.5 & 52.5 $\pm$ 0.7 & 13.1 $\pm$ 0.3 & 91.8 $\pm$ 0.5 & 76.4 $\pm$ 0.6 & 53.1 $\pm$ 0.4 & 17.3 $\pm$ 0.4 \\
DMGIattn  & 90.9 $\pm$ 0.4 & 70.2 $\pm$ 0.3 & 52.6 $\pm$ 0.3 & 9.2 $\pm$ 0.2 & 91.3 $\pm$ 0.4 & 75.2 $\pm$ 0.4 & 52.3 $\pm$ 0.5 & 17.1 $\pm$ 0.3 \\
HDMI  & 90.8 $\pm$ 0.4 & 69.5 $\pm$ 0.5 & 57.6 $\pm$ 0.4 & 14.5 $\pm$ 0.4 & 90.1 $\pm$ 0.4 & 73.1 $\pm$ 0.3 & 58.3 $\pm$ 0.3 & 20.3 $\pm$ 0.4 \\
HeCo & 88.4 $\pm$ 0.6 & 67.8 $\pm$ 0.8 & 50.9 $\pm$ 0.5 & 10.1 $\pm$ 0.6 & 89.2 $\pm$ 0.3 & 71.0 $\pm$ 0.7 & 58.4 $\pm$ 0.6 & 20.4 $\pm$ 0.5 \\
MCGC & 90.4 $\pm$ 0.5 & 69.0 $\pm$ 0.5 & 56.5 $\pm$ 0.3 & 14.9 $\pm$ 0.4 &\textbf{ 91.9 $\pm$ 0.2} & \textbf{76.5 $\pm$ 0.4} & 58.1 $\pm$ 0.4 &{ 20.2 $\pm$ 0.3} \\
CKD & 90.6 $\pm$ 0.4 & 69.3 $\pm$ 0.3 & 53.9 $\pm$ 0.3 & 13.8 $\pm$ 0.4 & 91.4 $\pm$ 0.4 & 75.9 $\pm$ 0.4 & 58.5 $\pm$ 0.6 & 20.6 $\pm$ 0.4 \\
RGRL&  90.7 $\pm$ 0.5 &  69.4 $\pm$  0.2 &   52.5  $\pm$ 0.5 &   13.3 $\pm$ 0.2 &   91.0 $\pm$ 0.3 &  74.5  $\pm$ 0.5 &  58.9 $\pm$ 0.3 &  20.8 $\pm$ 0.4 \\
\textbf{CoCoMG} &\textbf{92.3   $\pm$ 0.1}&\textbf{73.6   $\pm$ 0.1}&\textbf{59.0   $\pm$ 0.4 } &\textbf{ 17.4  $\pm$ 0.2}  & 91.3  $\pm$ 0.2 &  75.6 $\pm$  0.3&\textbf{ 64.0 $\pm$ 0.1}&\textbf{24.1 $\pm$ 0.1}\\

\bottomrule
\end{tabular}

\label{tabcluster}
\end{table*}

\begin{figure*}[h]
\centering
\includegraphics[scale=0.27]{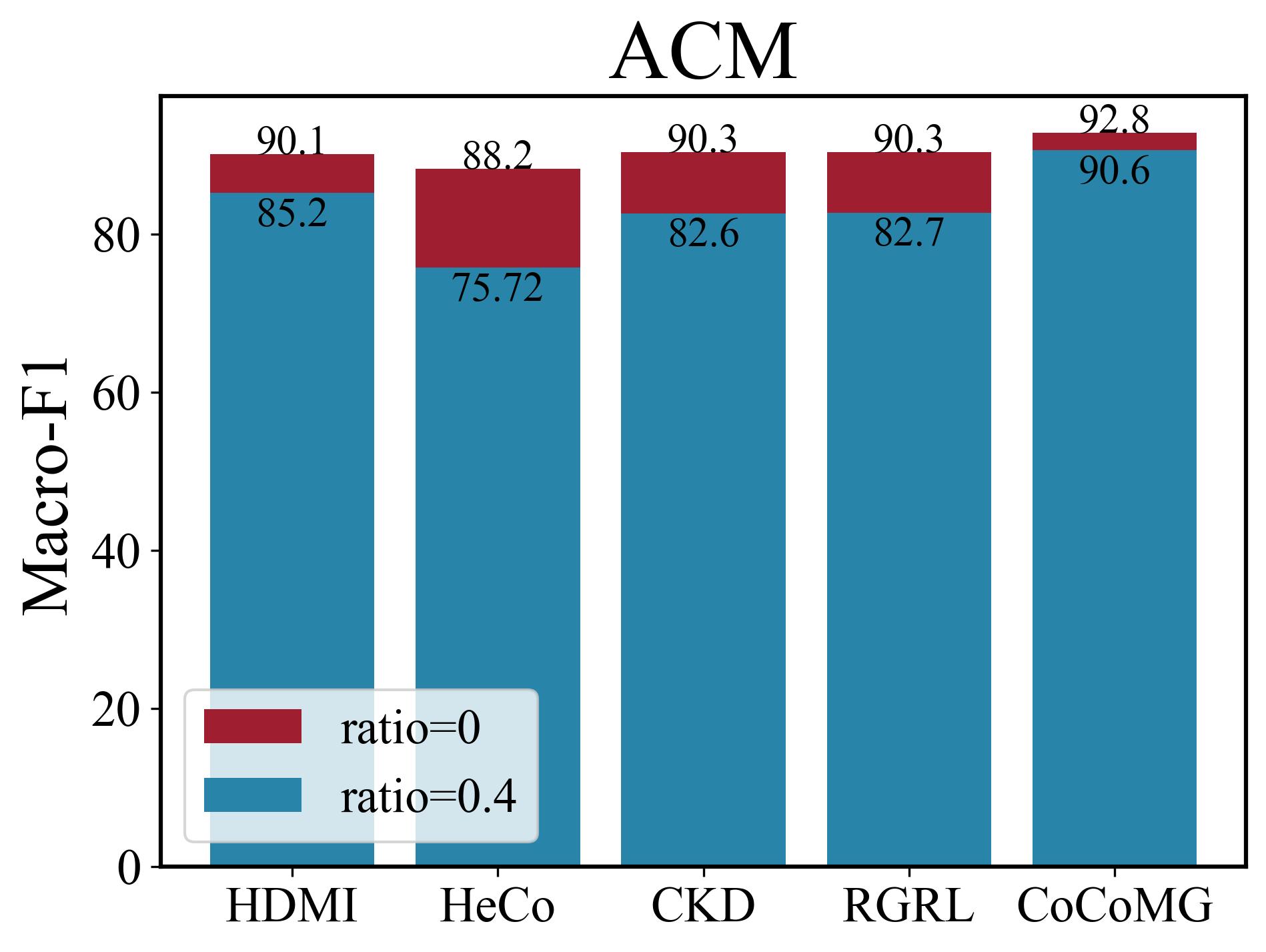}
\includegraphics[scale=0.27]{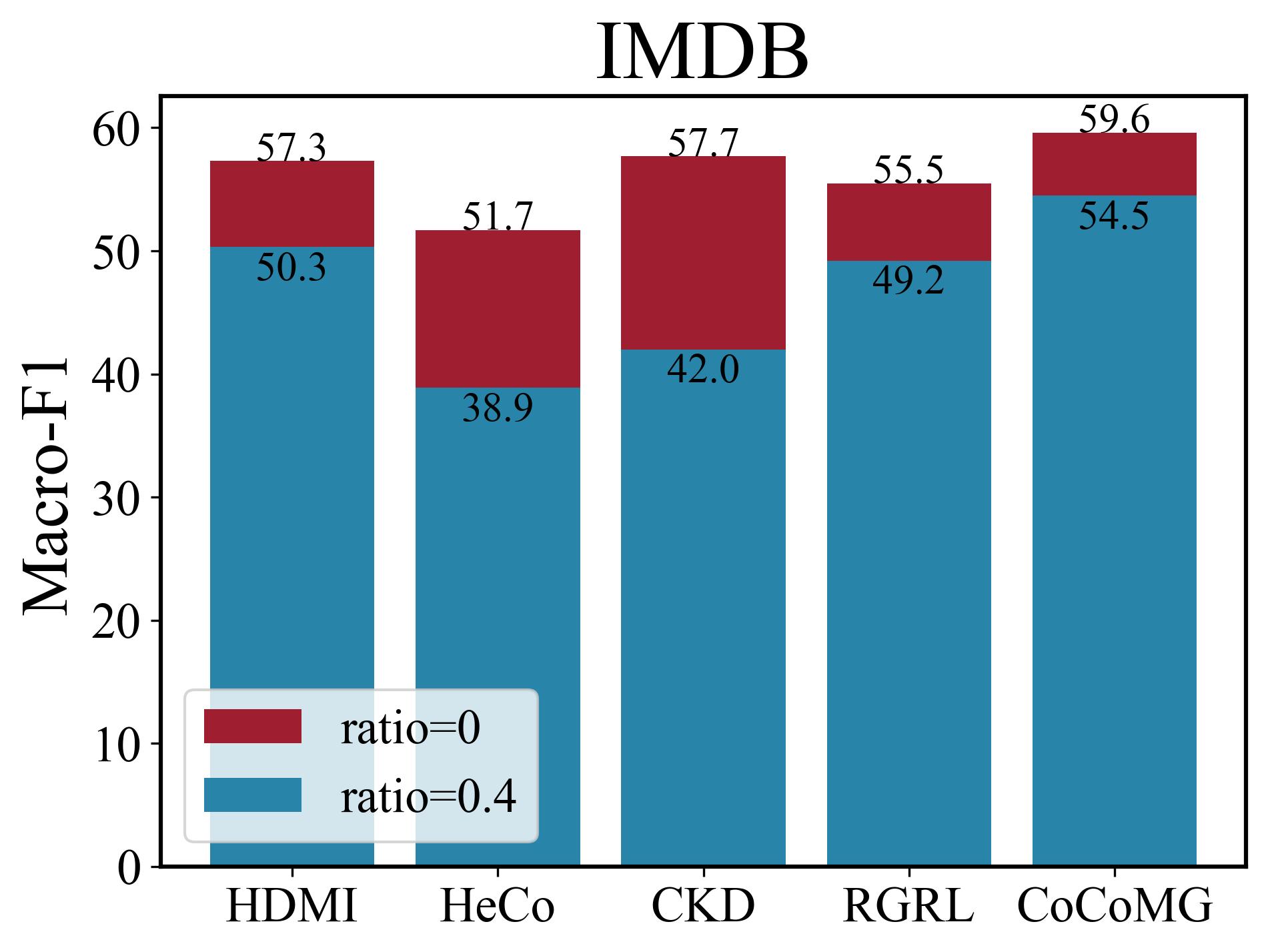}
\includegraphics[scale=0.27]{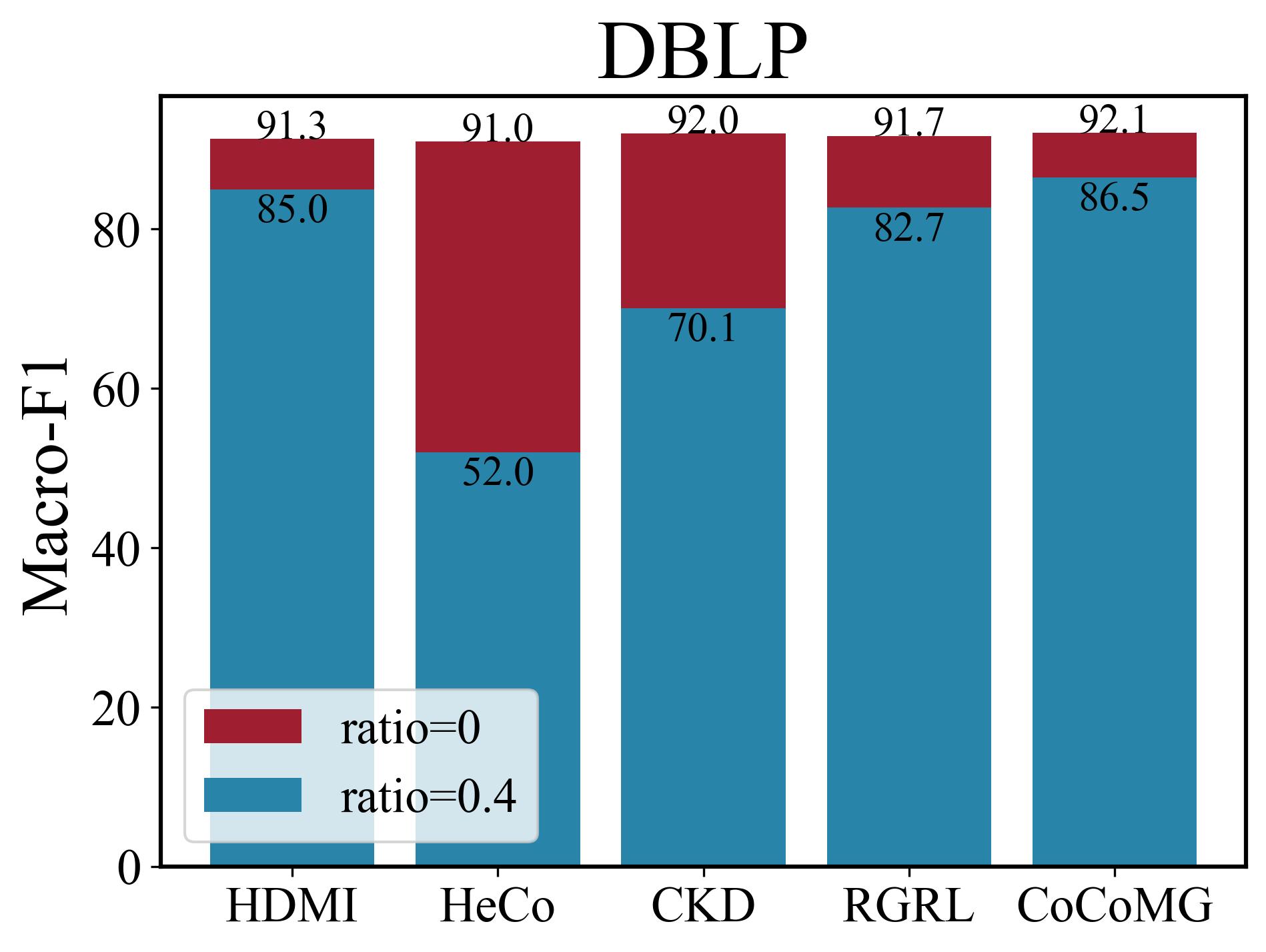}
\includegraphics[scale=0.27]{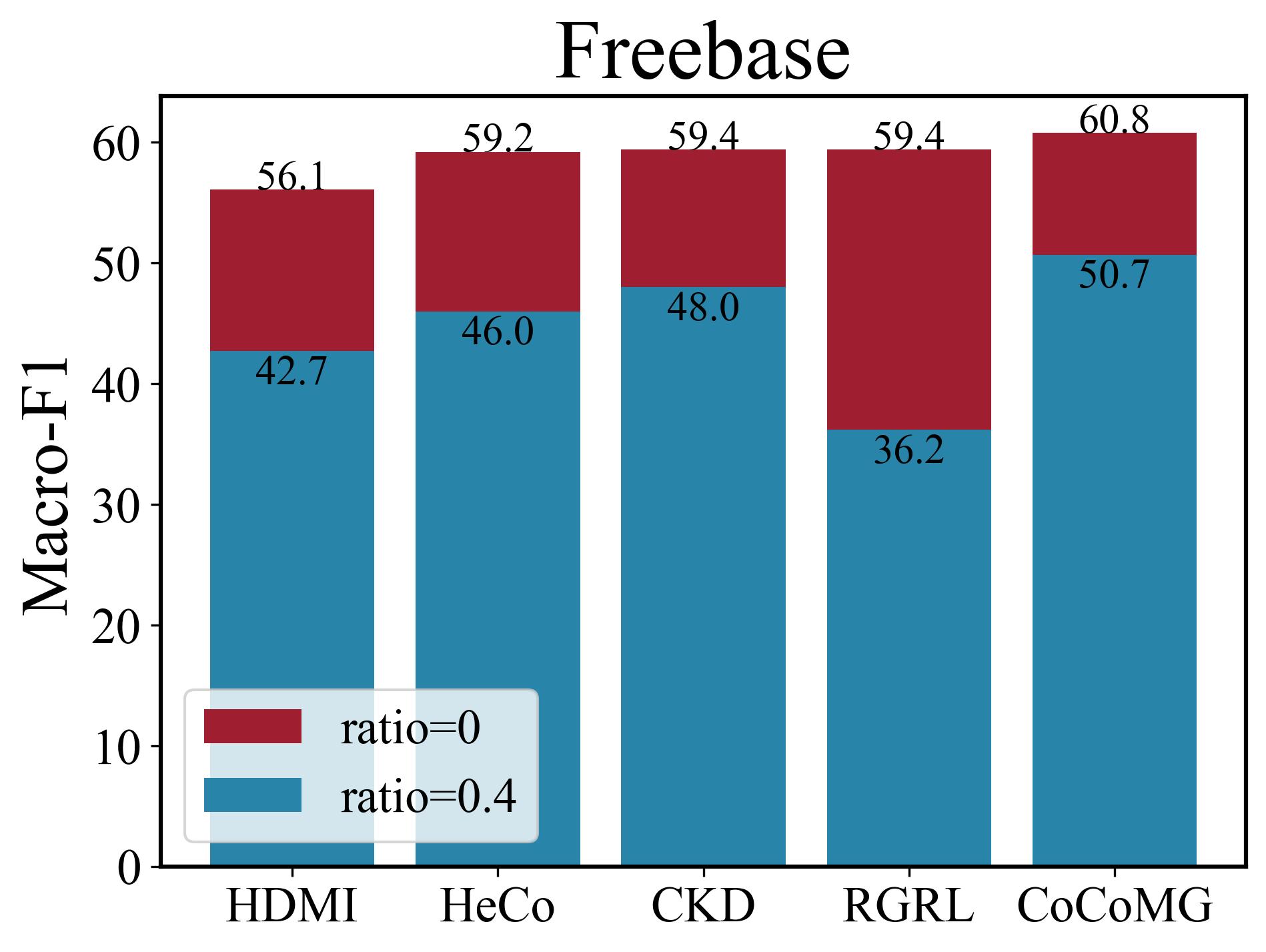}
\caption{Classification performance of our method and four UMGL methods for the out-of-sample extension on all multiplex graph datasets. The ``ratio'' indicates the percentage of unseen nodes in each dataset.}
\label{fig_sample}
\end{figure*}

\begin{figure*}[h]
\vspace{1mm}
\centering
\includegraphics[scale=0.27]{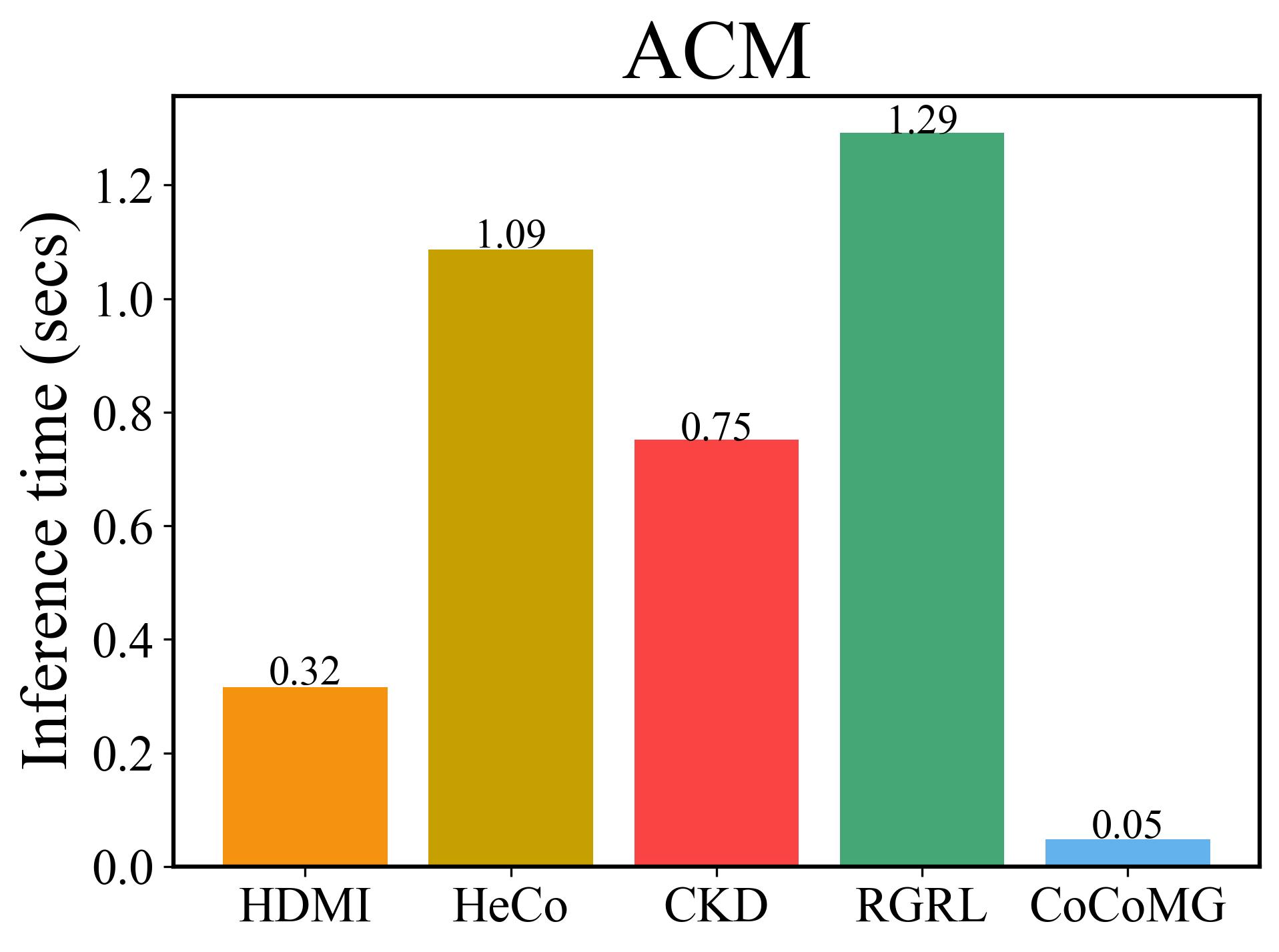}
\includegraphics[scale=0.27]{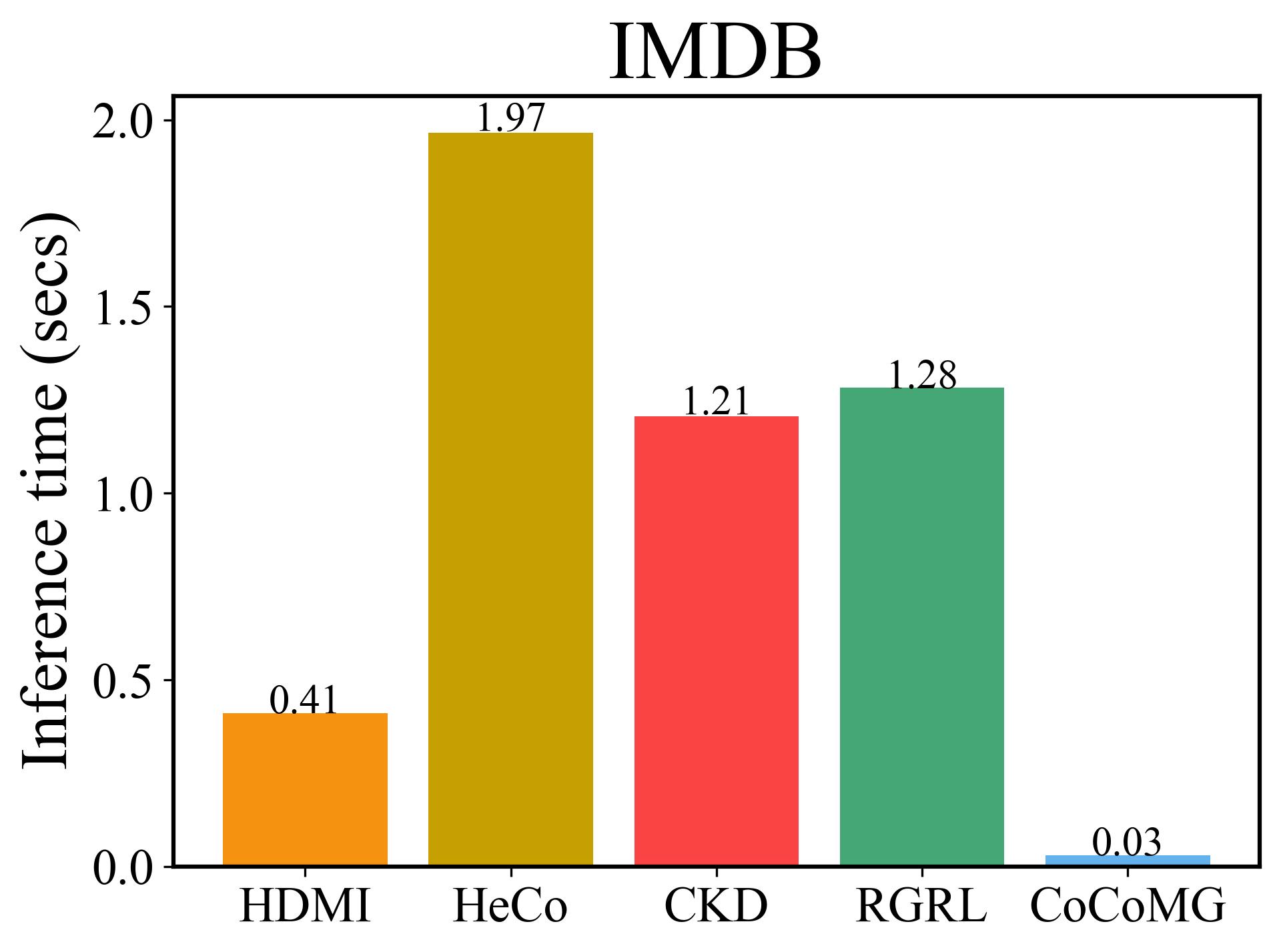}
\includegraphics[scale=0.27]{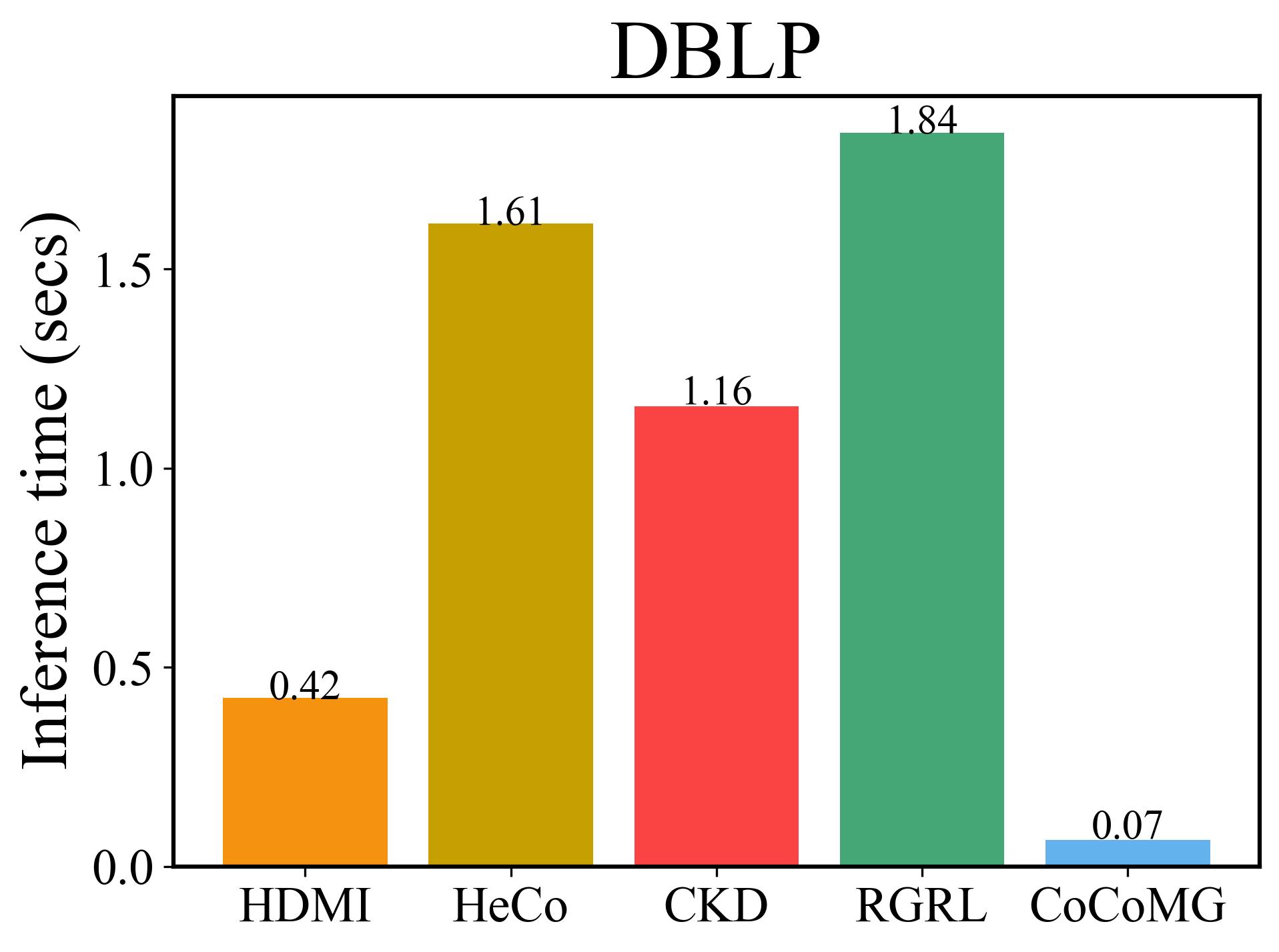}
\includegraphics[scale=0.27]{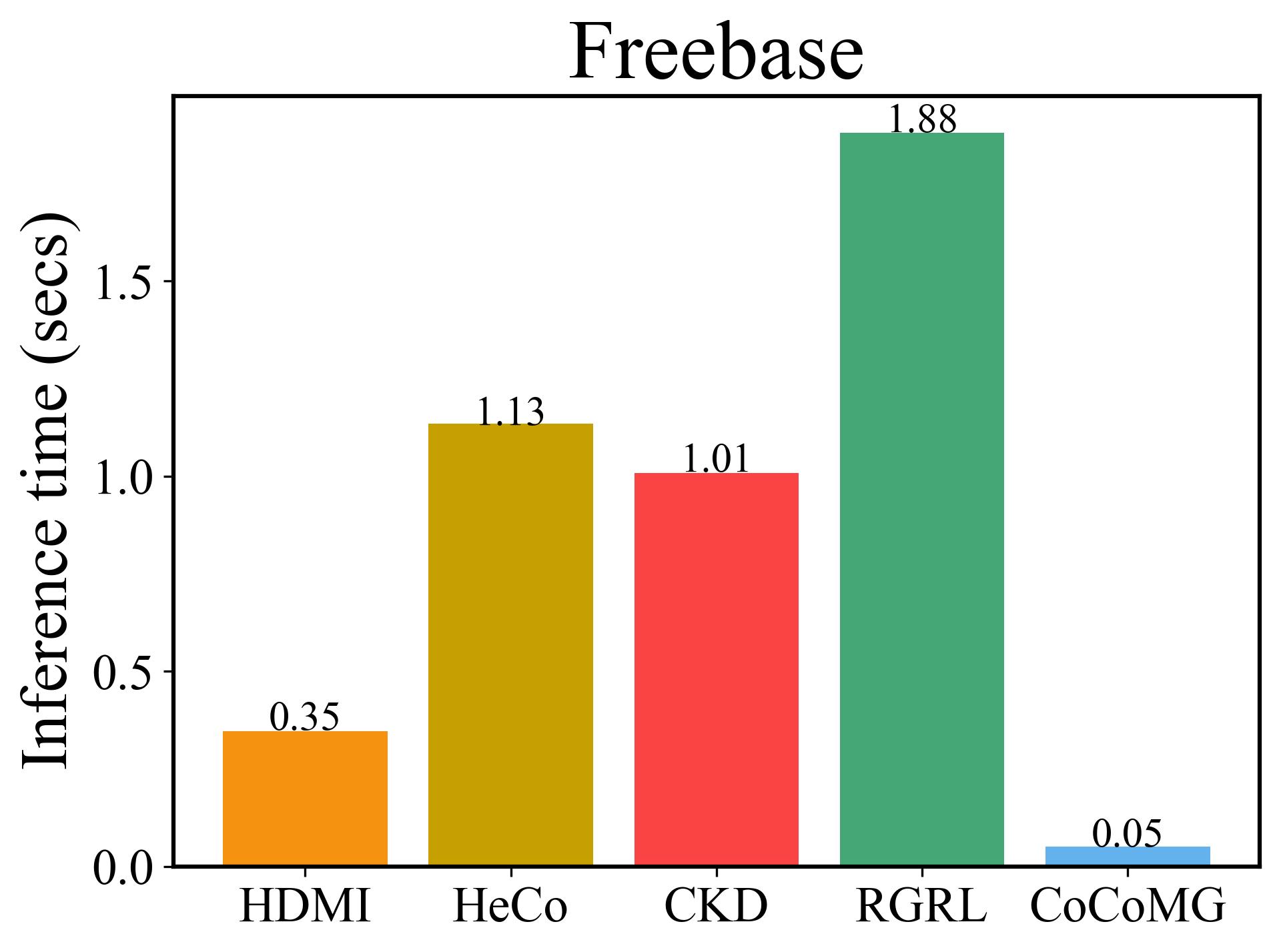}
\caption{Inference time (seconds) of \ours{} and four UMGL methods for the out-of-sample extension on all datasets.}
\label{Inference_time}
\end{figure*}

\begin{figure*}[h]
\centering
\includegraphics[scale=0.27]{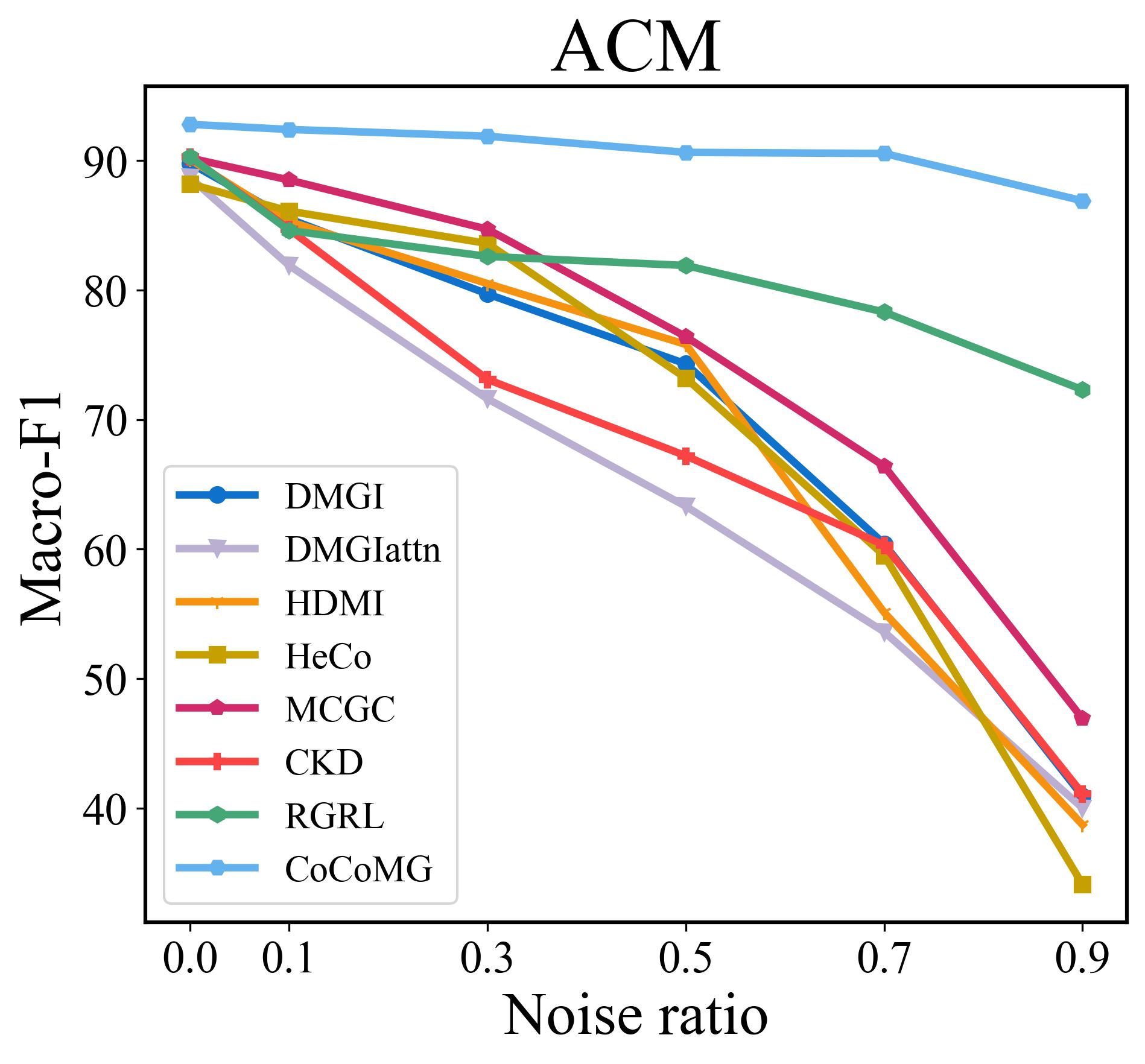}
\includegraphics[scale=0.27]{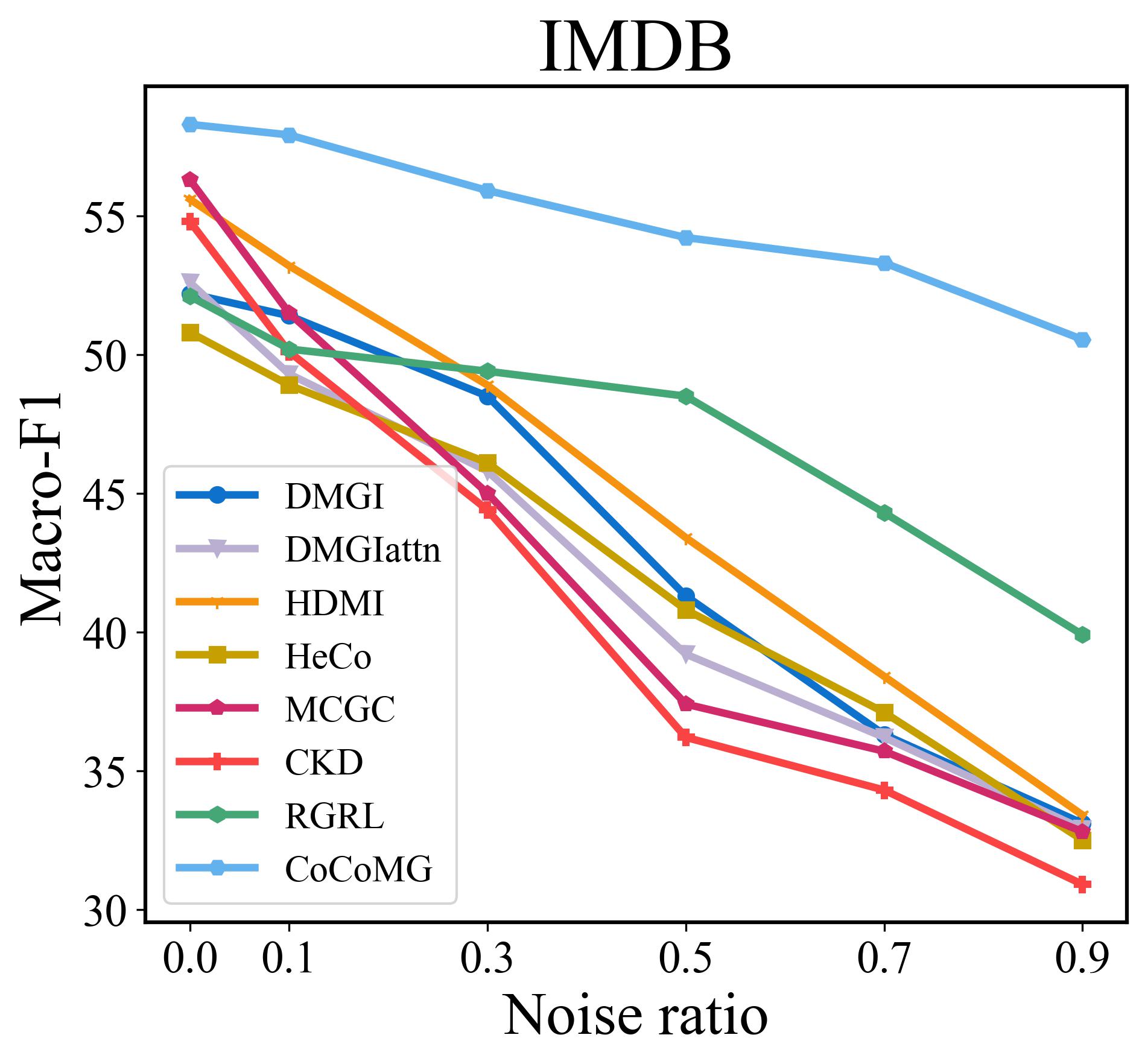}
\includegraphics[scale=0.27]{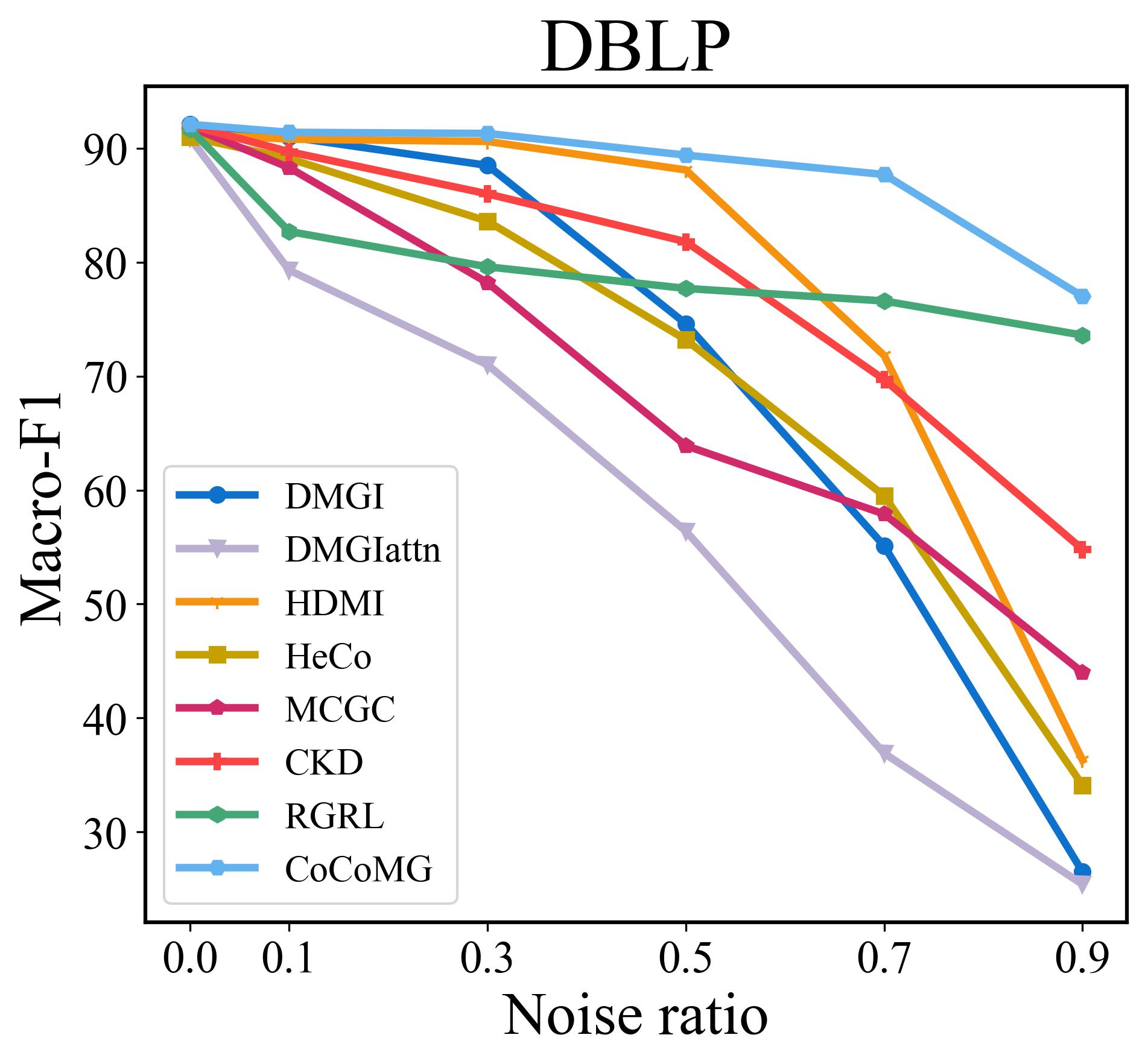}
\includegraphics[scale=0.27]{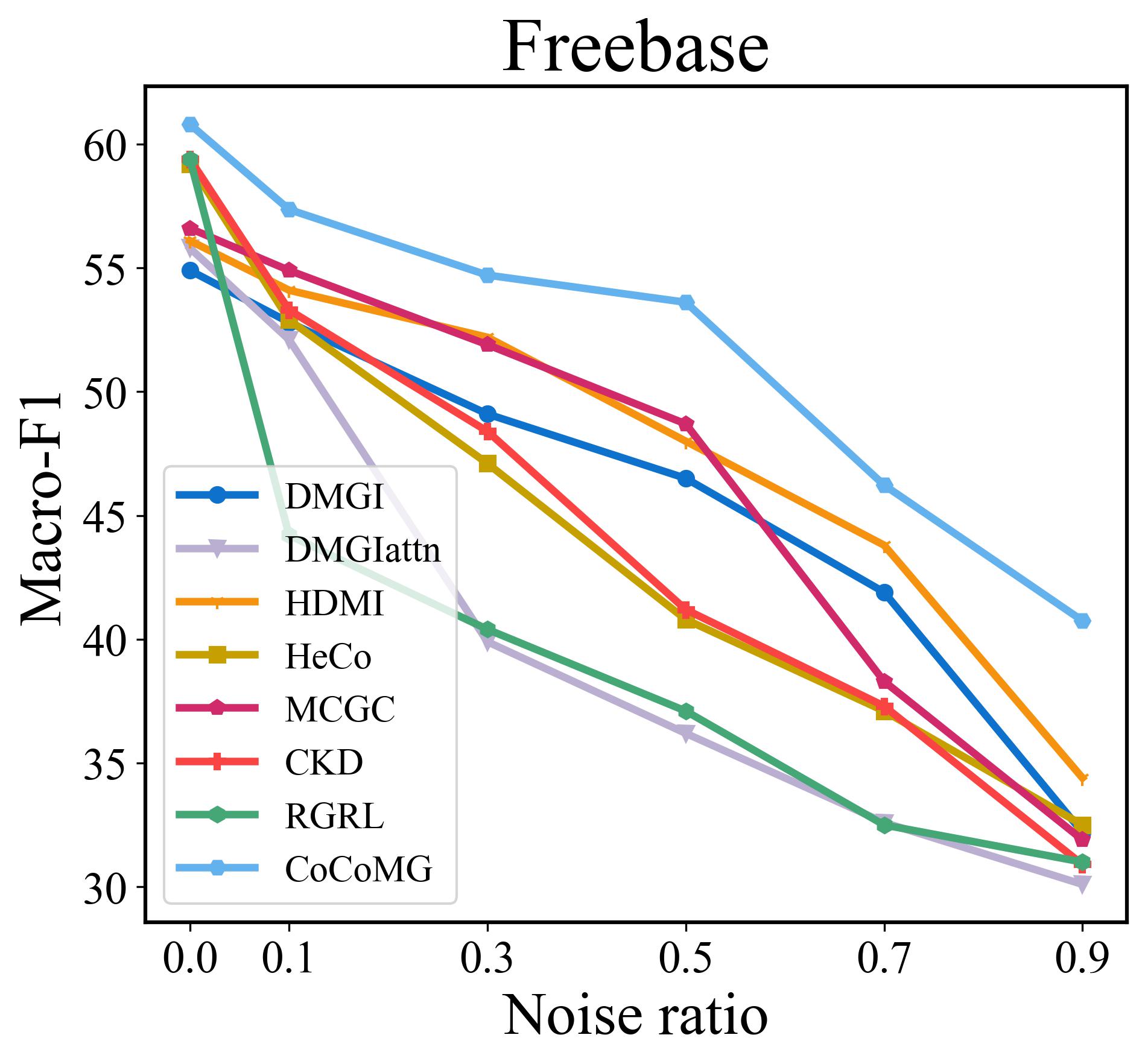}
\vspace{-2mm}
\caption{Classification performance of our method and all self-supervised multiplex graph representation methods under different noisy edges ratios $\eta$ on all multiplex graph datasets.}
\vspace{-1mm}
\label{fig_noise}
\end{figure*}

\section{Experiments} \label{sec_exp}

\subsection{Experimental Setup}

\subsubsection{Datasets}

We use four public benchmark datasets to evaluate the performance of our proposed method, including two citation multiplex graph networks (\ie ACM \cite{WangJSWYCY19} and DBLP \cite{DMGIParkK0Y20}), and two movie multiplex graph networks (\ie IMDB \cite{DMGIParkK0Y20} and Freebase \cite{jing2021hdmi}). We follow the public splitting settings \cite{jing2021hdmi, WangLHS21} in our experiments.

\subsubsection{Comparison methods} The comparison methods include four single-view graph learning methods and eight multiplex graph learning methods. The single-view graph learning methods consist of two semi-supervised methods, \ie GCN~\cite{kipf2017semisupervised}, and GAT~\cite{velickovic2018graph}, one traditional unsupervised learning method, \ie DeepWalk~\cite{perozzi2014deepwalk}, and one self-supervised learning method, \ie DGI~\cite{VelickovicFHLBH19}. The multiplex graph learning methods include one traditional unsupervised learning method,\ie MNE~\cite{ZhangQYS18}, and seven self-supervised learning methods, \ie DMGI~\cite{DMGIParkK0Y20}, DMGIattn~\cite{DMGIParkK0Y20}, HDMI~\cite{jing2021hdmi}, HeCo \cite{WangLHS21}, MCGC~\cite{pan2021multi}, CKD~\cite{wang2022collaborative}, and RGRL~\cite{lee2022relational}. To ensure a fair comparison, we evaluate single-view graph methods on multiplex graph datasets by training them independently on each graph and then concatenating their learned representations for downstream tasks.  The average values of 10 runs are reported.

\subsubsection{Evaluation protocol}
We follow the evaluation in previous works \cite{jing2021hdmi, wang2022collaborative} to conduct node classification and node clustering as semi-supervised and unsupervised downstream tasks, respectively. Specifically, we employ Macro-F1 and Micro-F1 to evaluate the performance of node classification.  Moreover, we employ Accuracy and Normalized Mutual Information (NMI) to evaluate the performance of node clustering. Additionally, to evaluate the robustness of our method and comparison methods, we randomly replace a certain ratio of edges in each graph with noisy edges (\ie random edges).

\subsection{Effectiveness Analysis} \label{sec_exp_main}

To evaluate the effectiveness of the proposed method, we report the performance of node classification (\ie Macro-F1 and Micro-F1) and node clustering (\ie Accuracy and NMI) of all methods in Table \ref{tabnode} and \ref{tabcluster}, respectively. Obviously, the proposed method obtains the highest effectiveness on both node classification and node clustering tasks.

First, compared to the single graph learning methods (\ie DeepWalk, GCN, GAT, and DGI), the proposed \ours{} outperforms them by significant margins. For example, \ours{} achieves an average improvement of 11.4\% and 29.7\% in terms of classification and clustering tasks, respectively, compared to the best-performing unsupervised single graph method DGI. Moreover, most multiplex graph learning methods have achieved better performance than single graph methods on both node classification and clustering tasks
This suggests that multiplex graph learning methods are more effective than single-view graph methods, as they can explore the complementary and consistent information among different graphs to learn more discriminative node representations.


Second, compared to the multiplex graph learning methods, the proposed \ours{} achieved the best performance, followed by CKD, MCGC, RGRL, HDMI, HeCo, DMGI, DMGIattn, and MNE. 
The reason can be attributed to  that the proposed method can effectively extract complementary and consistent information while reducing the impact of noise.

\subsection{Out-of-sample Analysis} \label{sec_exp_out}
In this section, we compare the effectiveness and efficiency of the proposed method with previous UMGL methods when tackling the out-of-sample issue. Note that, the comparison method, such as DMGI \cite{DMGIParkK0Y20}, DMGIattn \cite{DMGIParkK0Y20}, and MCGC \cite{pan2021multi}, cannot handle the out-of-sample issue as they need to retrain the whole model to infer the representation of the unseen nodes. Thus, we only compare our method with HDMI \cite{jing2021hdmi}, HeCo \cite{WangLHS21}, CKD \cite{wang2022collaborative}, and RGRL \cite{lee2022relational} on all multiplex graph datasets. To do this, we first randomly split (repeated 5 times with random seeds) the unseen nodes from all nodes in the dataset with different ratios (\ie 0 and 0.4) for each dataset. We then train the model for each method on the in-sample nodes with their local graph structures. After the training is completed, we use the optimized model to infer the representation of unseen nodes and evaluate the inference time and the performance on the node classification task.

The performance results are shown in Figure \ref{fig_sample}. Firstly, in terms of horizontal comparison (\ie blue bars of all methods), the proposed method achieves the best performance on all datasets compared to the comparison methods, followed by HDMI, RGRL, CKD, and HeCo. Secondly, in terms of vertical comparison (\ie the performance gap between red bars and blue bars), we observe that the performance drop of our method compared to the comparison methods is the least significant. The reason can be that previous methods only preserve the graph structure without learning an inference model that can map the input to the graph structure. As a result, these methods lack the inference ability when facing unseen nodes and their local graph structures. In contrast, our method learns an inference model that can implicitly map features to the graph structure, allowing it to effectively infer the representations of unseen nodes by implicitly mapping them to the graph structures among the seen nodes.

The inference times of our proposed method and comparison methods are presented in Figure \ref{Inference_time}, demonstrating the superior efficiency of our approach. Across all multiplex graph datasets, our method is on average 33.1× and 8.3× faster than the slowest method RGRL and the fastest method HDMI, respectively. This can be attributed to the fact that our method only utilizes MLP encoders for representation inference. These results are promising, as they indicate that our proposed method is capable of effectively tackling the out-of-sample issue while achieving superior efficiency.


\begin{table*}[ht]
\centering
\caption{Classification performance (\ie Macro-F1 and Micro-F1) of each constraint in \ours{} on all mutiplex graph datasets.}
{
\centering
\begin{tabular}{cccccccccc}
\toprule

~&~&\multicolumn{2}{c}{\textbf{ACM}}& \multicolumn{2}{c}{\textbf{IMDB}}& \multicolumn{2}{c}{\textbf{DBLP}}& \multicolumn{2}{c}{\textbf{Freebase}} \\
\cmidrule(r){3-4} \cmidrule(r){5-6} \cmidrule(r){7-8} \cmidrule(r){9-10}

$\mathcal{L}_{LP}$&$\mathcal{L}_{CCA}$&Macro-F1&Micro-F1&Macro-F1&Micro-F1&Macro-F1&Micro-F1&Macro-F1&Micro-F1\\

\midrule

$-$&$\surd$& 86.8 $\pm$ 0.8  & 86.9 $\pm$ 0.7 &51.0 $\pm$ 0.4 &51.6 $\pm$ 0.8 &53.2 $\pm$ 0.2  &77.9 $\pm$ 0.2 & 35.3 $\pm$ 0.3 & 42.7 $\pm$ 0.8 \\

$\surd$&$-$&92.1 $\pm$ 0.2 &92.1 $\pm$ 0.1  &51.5 $\pm$ 0.6 &52.1 $\pm$ 0.8 &90.5 $\pm$ 0.4 &91.5 $\pm$ 0.4 &59.0 $\pm$ 0.5 &62.4 $\pm$ 0.1 \\

$\surd$&$\surd$ &\textbf{92.8 $\pm$ 0.1} &\textbf{92.7 $\pm$ 0.1} &\textbf{58.3 $\pm$ 0.4} &\textbf{59.6 $\pm$ 0.4} &\textbf{92.1 $\pm$ 0.1}  &\textbf{92.9 $\pm$ 0.1}& \textbf{60.8 $\pm$ 0.1} & \textbf{63.8 $\pm$ 0.9} \\

\bottomrule
\end{tabular}
}
\label{tabab}
\end{table*}

\begin{figure*}[h]
\centering
\subfigure[w/o $\mathcal{L}_{LP}$(SIL=0.42)]{
    \label{fig_visualization_a}
    \includegraphics[scale=0.3]{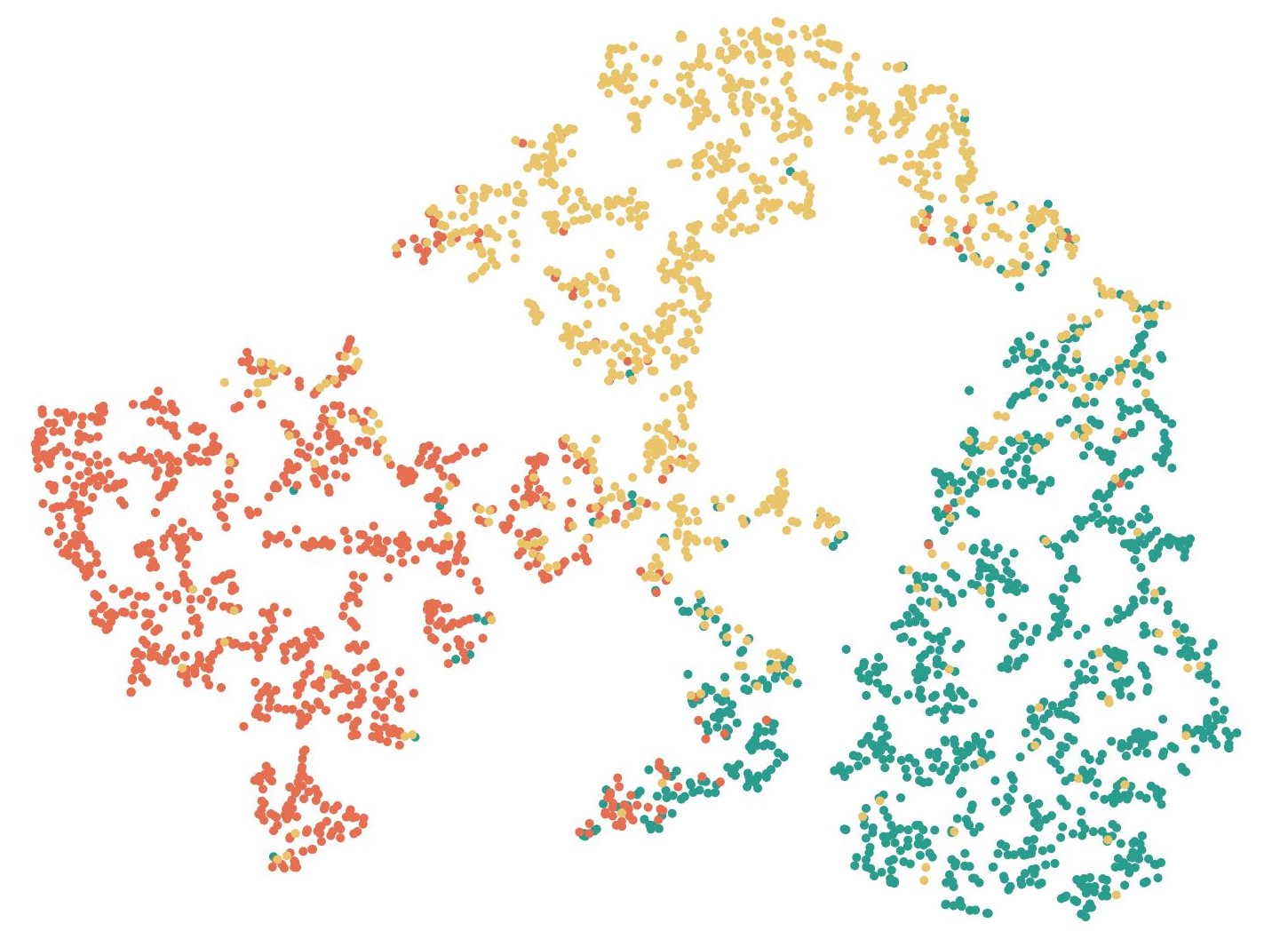}
}
\hspace{12.5mm}
\subfigure[w/o $\mathcal{L}_{CCA}$(SIL=0.53)]{
    \label{fig_visualization_b}
    \includegraphics[scale=0.3]{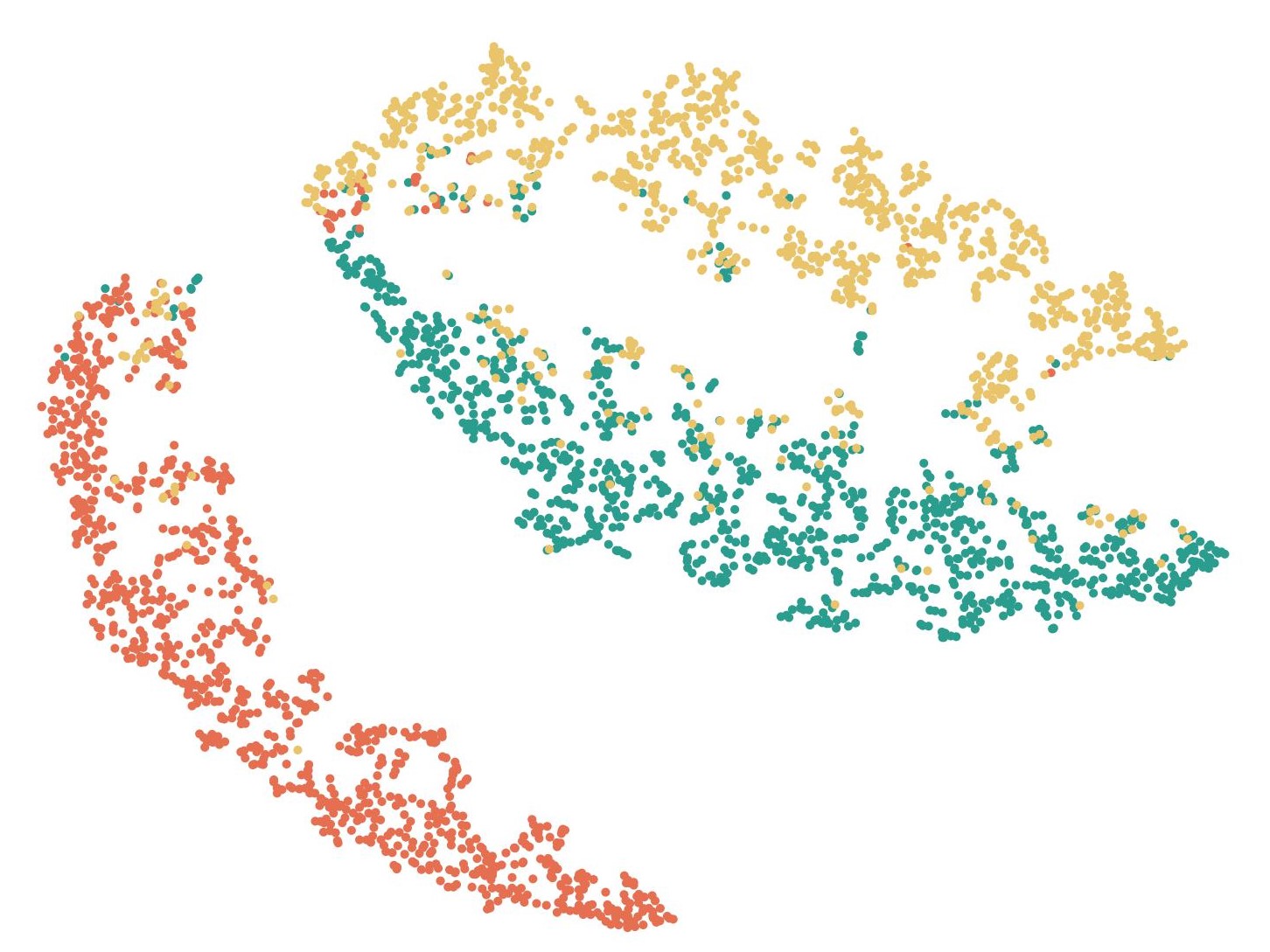}
}
\hspace{12.5mm}
\subfigure[with $\mathcal{J}$(SIL=0.74)]{
    \label{fig_visualization_c}
    \includegraphics[scale=0.3]{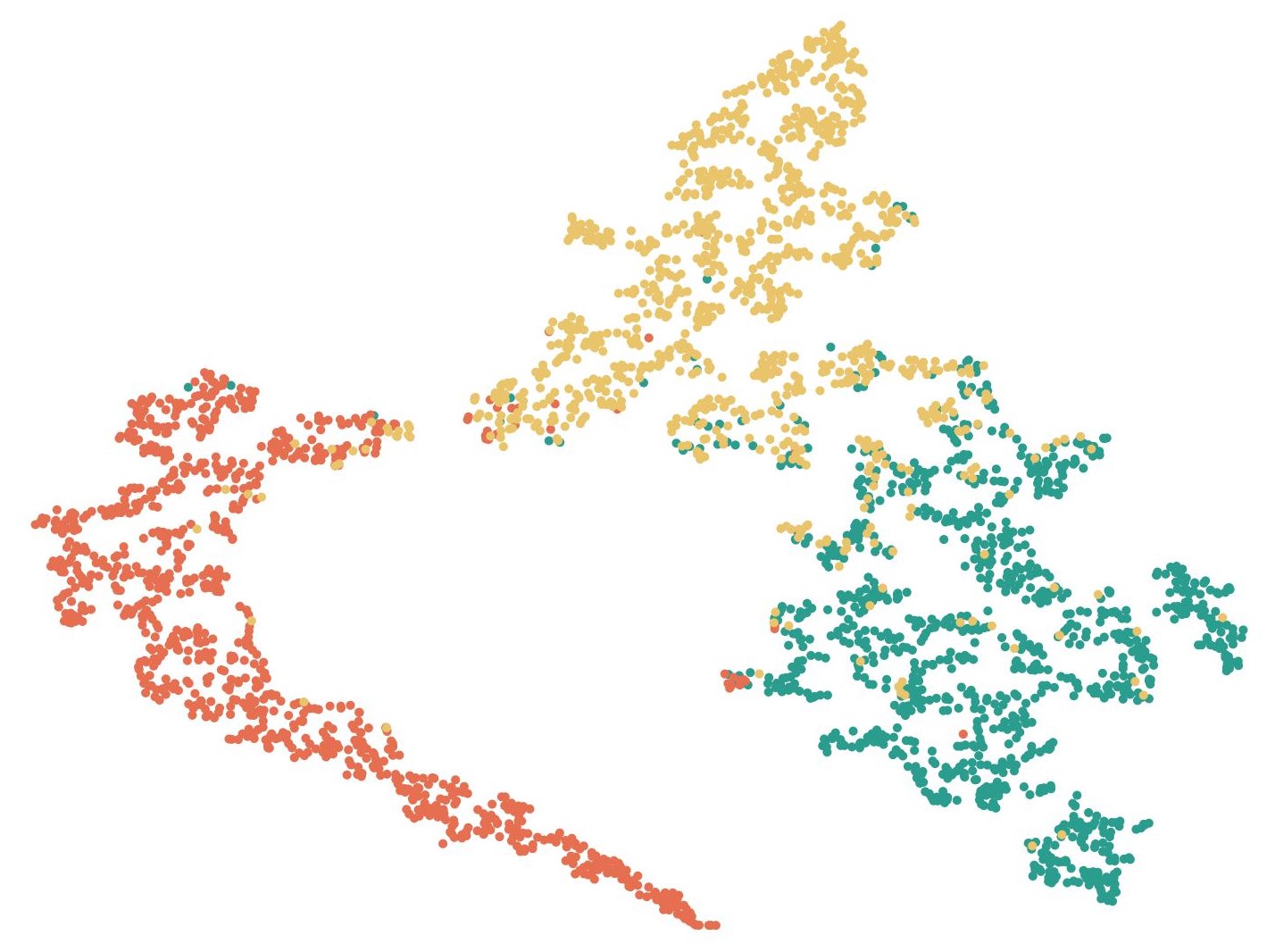}
}

\caption{Visualization plotted by $t$-SNE and the corresponding silhouette scores (SIL) of node representations for our method on Dataset ACM, where different colors represent different classes of the nodes. (a) CoCoMG without $\mathcal{L}_{LP}$; (b) CoCoMG without $\mathcal{L}_{CCA}$; (c) CoCoMG with two constraints.}
\label{fig_visualization}
\end{figure*}

\subsection{Robustness Analysis} \label{sec_exp_robust}

To verify the robustness of the proposed method, we report the performance of all self-supervised multiplex graph learning methods on the node classification task under varying ratios of noisy edges (random edges) ranging from 0.1 to 0.9. These results are presented in Figure \ref{fig_noise}.

From Figure \ref{fig_noise}, we have the observations as follows. Firstly, the proposed method consistently achieves the best performance on all datasets under different noise ratios, confirming our claim that \ours{} obtains satisfying prediction performance under noise. 
Secondly, as the noise ratio increases, the performance of all methods decreases, while the proposed method exhibits the smallest decrease. For instance, on Dataset ACM, the proposed method is minimally affected in terms of performance when the noise ratio is below 0.7. Even when the noise ratio reaches 0.9, the performance only decreases by approximately 6\%, while other methods show a decrease of over 40\%. The reasons behind these results are twofold. Firstly, we only use the MLP encoders to obtain node representations, which means noise cannot be directly aggregated into the node representations through incorrect edges. Secondly, we use Eq. (\ref{eq_loss_LCCA}) to ensure the consistency of the representations between different views, which reduces the noise caused by Eq. (\ref{eq_loss_LP}).

\subsection{Ablation Study} \label{sec_exp_ab}

The proposed method employs two constraints (\ie $\mathcal{L}_{LP}$ in Eq. (\ref{eq_loss_LP}) and $\mathcal{L}_{CCA}$ in Eq. (\ref{eq_loss_LCCA})) to extract complementary and consistent information, respectively. To verify the effectiveness of each constraint of the proposed method, we investigate the performance(\ie Macro-F1 and Micro-F1) of each constraint on the node classification task, and report the results in Table \ref{tabab}. Generally, we observe that the proposed method with two constraints achieves the best performance. Conversely, when we only implement an individual constraint (\ie either $\mathcal{L}_{LP}$ or $\mathcal{L}_{CCA}$), the performance is inferior. Moreover, as shown in Figure \ref{fig_visualization}, we utilize $t$-SNE \cite{van2008visualizing} to visualize the node representations learned with different constraints on the ACM dataset. We observed that the representations of nodes belonging to the same class, learned with both constraints, are closer to each other than the representations learned with only one constraint.
Consequently, the experiments confirm that both constraints play a crucial role in the effectiveness of \ours{}.

\subsection{Over-smooth Analysis} 
\label{app_additional}

Previous UMGL methods have relied heavily on GCN to obtain node embeddings. However, as the number of layers in a GCN increases, the performance of the model tends to decrease, which is known as the over-smooth problem. In fact, modern deep neural networks typically require multiple layers to effectively extract features and learn complex patterns, but in graph neural networks, using too many graph convolutional layers can lead to the over-smooth phenomenon, which limits the number of layers in a GCN. To some extent, this presents a paradoxical situation in which the depth of the GCN is both necessary and limited.

\begin{figure}[!h]
\centering
\includegraphics[scale=0.5]{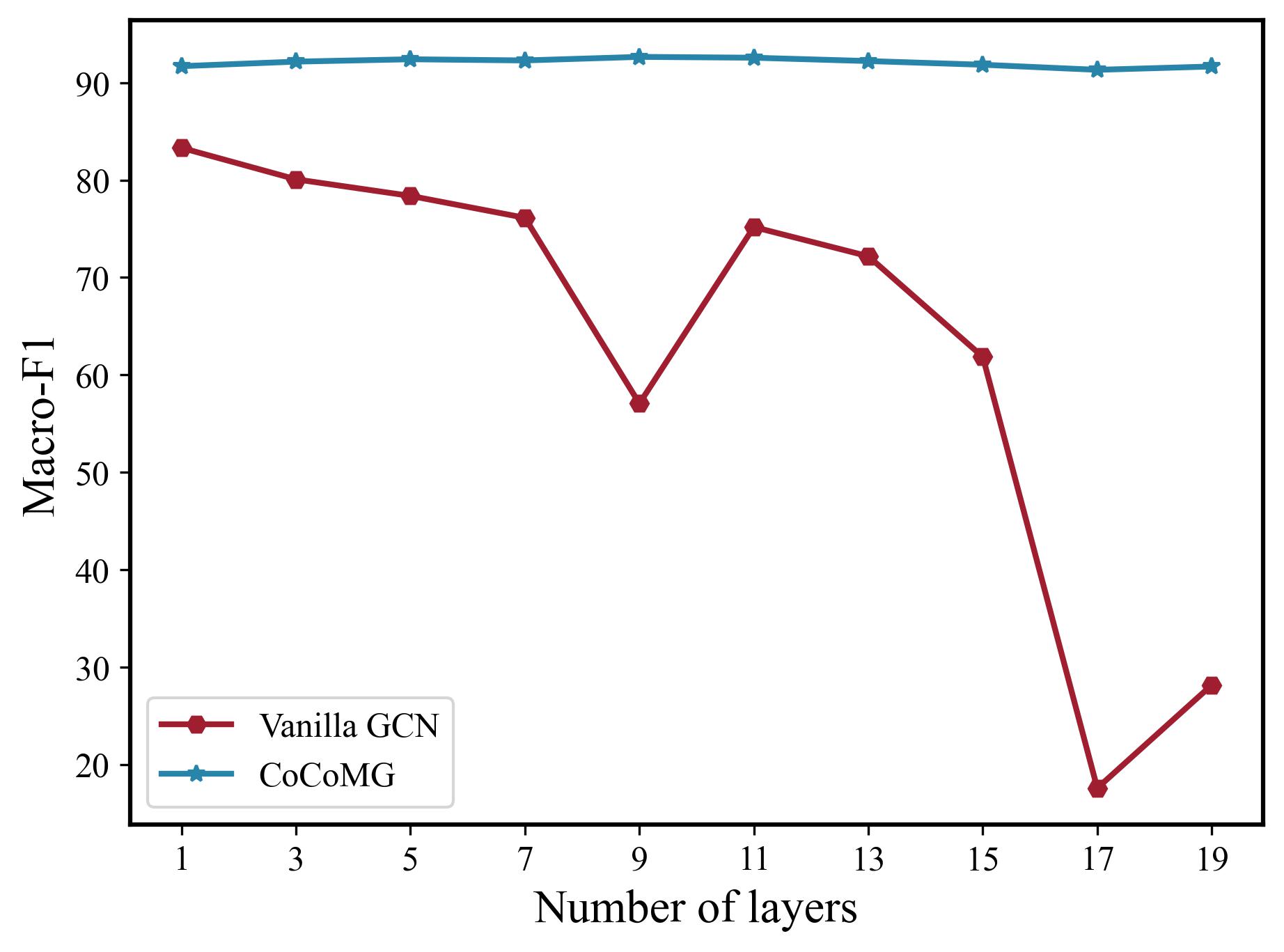}
\caption{Classification results of our method and vanilla GCN under different number of layers on Dataset ACM}
\label{smooth}
\end{figure}

In this section, we analyze the effectiveness of the proposed method when dealing with the over-smooth problem. For this purpose, we evaluate the effectiveness of our method and vanilla GCN (\ie using GCN as encoder) with varying numbers of layers ranging from $1$ to $19$  on Dataset ACM, and the results are presented in Figure \ref{smooth}. Overall, the performance of our method shows a slight increase and remains stable with the number of layers increasing, while vanilla GCN's performance drops sharply. Specifically, our model's performance improves as the number of layers increases from $1$ to $11$, and the performance of our model remains stable with the increase in the number of layers from $11$ to $19$. In contrast, for vanilla GCN, the performance drops significantly with the number of layers from $1$ to $19$, with a drop of over 60\%. The reason might be as follows. For our model, as the number of layers increases, the model can learn more abstract node embeddings, and thus the performance of the model increases gradually. For vanilla GCN, however, the message passing mechanism causes the node embeddings to become more and more similar as the layers increase, significantly degrading performance. Based on this observation, we can conclude that our method can effectively address the over-smoothing problem.

\subsection{Parameter Sensitivity} \label{sec_exp_pa}
\begin{figure}[h]
\vspace{-1mm}
\centering
\includegraphics[scale=0.38]{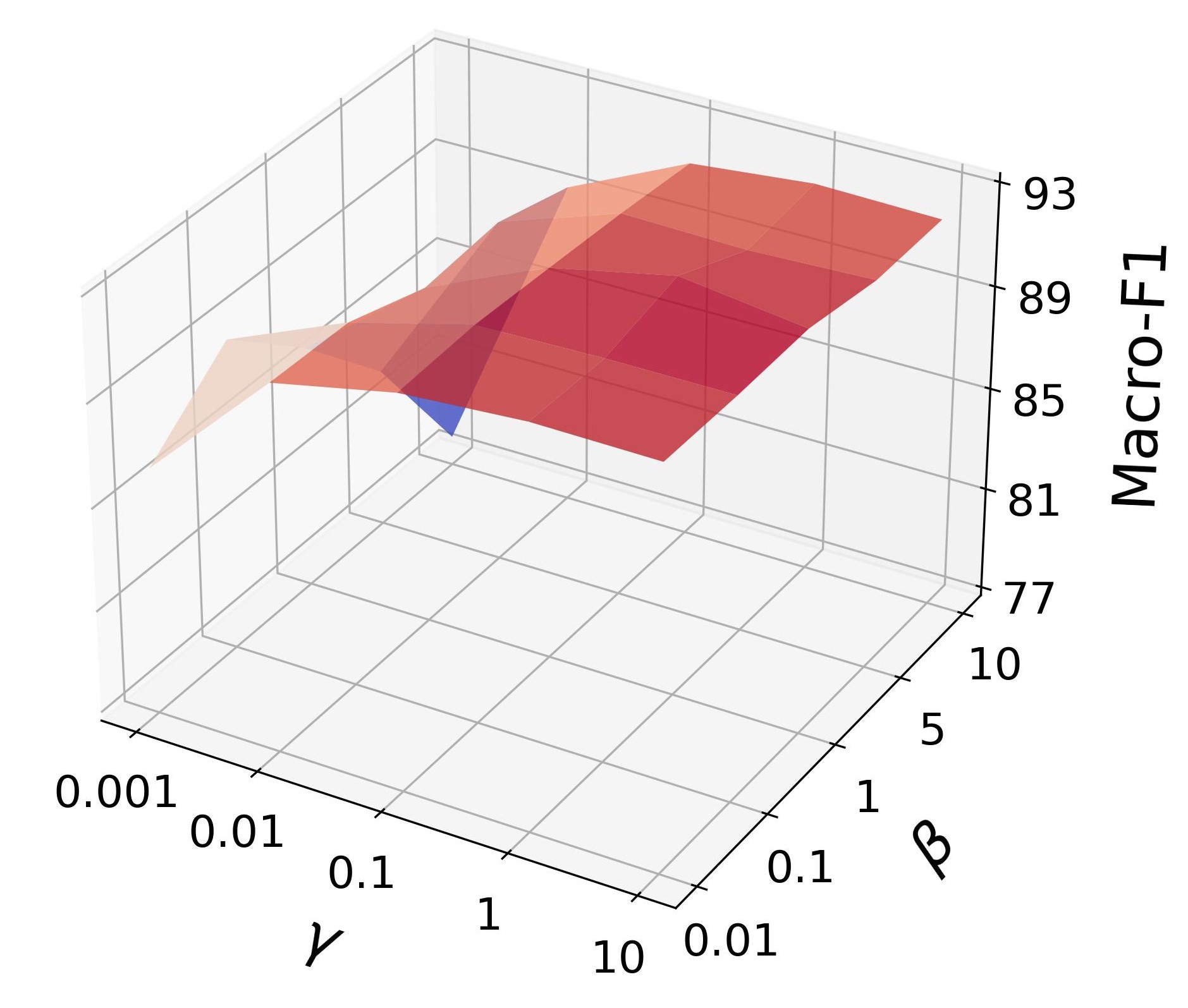}
\caption{Classification results of our method at different parameter settings (\ie $\gamma$ and $\beta$) on Dataset ACM.}
\label{fig_hyper}
\vspace{-2mm}
\end{figure}

The hyper-parameter of \ours{} includes the trade-off $\gamma$ in Eq. (\ref{eq_loss_LCCA}) and $\beta$ in Eq. (\ref{eq_loss}), and Figure \ref{fig_hyper} shows the effectiveness by traversing $\gamma$ and $\beta$ on Dataset ACM. The results indicate that $\beta$ is relatively insensitive in the range of 0.01 to 10, while $\gamma$ exhibits little sensitivity to performance. These results suggest that \ours{} can achieve near-perfect performance when  the value of $\gamma$ is in a reasonable range (\eg $\gamma > 0.1$).

\section{Conclusion} \label{sec_conclusion}
In this paper, we investigated two pervasive but challenging issues of existing UMGL methods, \ie the out-of-sample issue and the noise issue. To tackle these issues, we proposed \ours{},  that only employs multiple MLP encoders with two constraints to effectively and efficiently extract complementary and consistent information for conducting unsupervised multiple graph learning. 
Comprehensive experiments demonstrate that \ours{} achieves state-of-the-art performance in terms of both effectiveness and efficiency, as well as confirm our claim that \ours{} can tackle those two issues.



\normalem{

}

\end{document}